\documentclass[10pt,twocolumn,letterpaper]{article}
\usepackage{cvpr}
\usepackage{times}
\usepackage{epsfig}
\usepackage{graphicx}
\usepackage{amsmath}
\usepackage{amssymb}
\usepackage{booktabs}   

\usepackage{adjustbox}
\usepackage{subcaption}
\usepackage{color}
\usepackage{graphicx}
\usepackage{subcaption}
\usepackage{xspace}
\usepackage{amssymb}
\usepackage{multirow}
\usepackage{multicol}
\usepackage{algorithm}
\usepackage{algpseudocode}
\usepackage{colortbl}
\usepackage{float}
\usepackage{diagbox}
\usepackage{makecell}

\usepackage[pagebackref=true,breaklinks=true,letterpaper=true,colorlinks,bookmarks=false]{hyperref}

\newcommand{\minisection}[1]{\noindent{\bf #1}}
\newcommand{\ra}[1]{\renewcommand{\arraystretch}{#1}}
\newcommand\ver[1]{\small\rotatebox{90}{#1}}

\def\cvprPaperID{6723}

\cvprfinalcopy

\ifcvprfinal\fi
\begin{document}
\title{BDD100K: A Diverse Driving Dataset for Heterogeneous Multitask Learning}

\author{Fisher Yu$^1$\quad Haofeng Chen$^1$\quad Xin Wang$^1$\quad Wenqi Xian$^{2}\thanks{Work done at UC Berkeley.}$\quad Yingying Chen$^{1}$\quad \\
Fangchen Liu$^{3*}$\quad Vashisht Madhavan$^{4*}$\quad Trevor Darrell$^1$\\[2mm] 
$^1$UC Berkeley\quad $^2$Cornell University \quad $^3$UC San Diego \quad $^4$Element, Inc.
}

\maketitle

\begin{abstract}

Datasets drive vision progress, yet existing driving datasets are impoverished in terms of visual content and supported tasks to study multitask learning for autonomous driving. Researchers are usually constrained to study a small set of  problems on one dataset, while real-world computer vision applications require performing tasks of various complexities. We construct BDD100K~\footnote{The data is available at \url{https://bdd-data.berkeley.edu}}, the largest driving video dataset  with 100K videos and 10 tasks to evaluate the exciting progress of image recognition algorithms on autonomous driving. The dataset possesses geographic, environmental, and weather diversity, which is useful for training models that are less likely to be surprised by new conditions. Based on this diverse dataset, we build a benchmark for heterogeneous multitask learning and study how to solve the tasks together. Our experiments  show that special training strategies are needed for existing models to perform such heterogeneous tasks. BDD100K opens the door for future studies in this important venue.

\end{abstract}

\section{Introduction}
Diverse, large-scale annotated visual datasets, such as ImageNet~\cite{imagenet} and COCO~\cite{mscoco}, have been the driving force behind recent advances in supervised learning tasks in computer vision. Typical deep learning models can require millions of training examples to achieve state-of-the-art performance for a task~\cite{he2016deep,ren2015faster,he2017mask}.

\begin{figure*}[htp]

\centering
\bgroup
\def\arraystretch{0.35}
\begin{tabular}{c@{\hspace{1pt}}c@{\hspace{1pt}}c@{\hspace{1pt}}c}
    \includegraphics[width=0.24\linewidth]{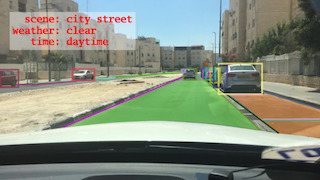} & 
    \includegraphics[width=0.24\linewidth]{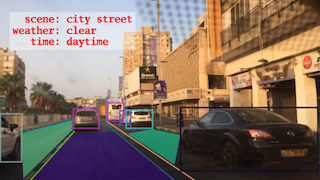} & 
    \includegraphics[width=0.24\linewidth]{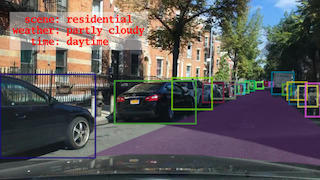} &
    \includegraphics[width=0.24\linewidth]{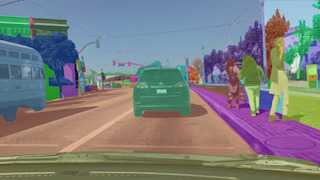} \\
    \includegraphics[width=0.24\linewidth]{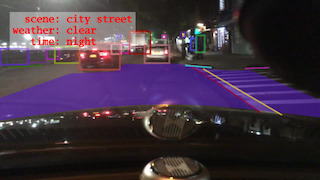} & 
    \includegraphics[width=0.24\linewidth]{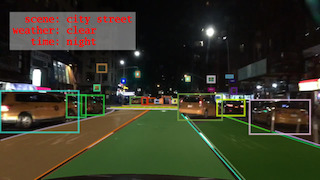} &
    \includegraphics[width=0.24\linewidth]{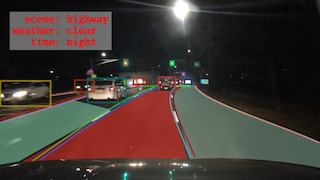} &
    \includegraphics[width=0.24\linewidth]{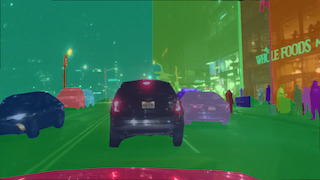} \\
    \includegraphics[width=0.24\linewidth]{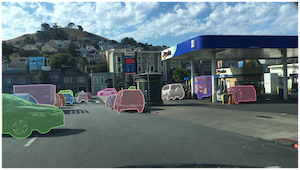} & 
    \includegraphics[width=0.24\linewidth]{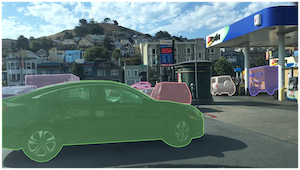} &
    \includegraphics[width=0.24\linewidth]{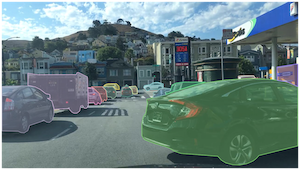} &
    \includegraphics[width=0.24\linewidth]{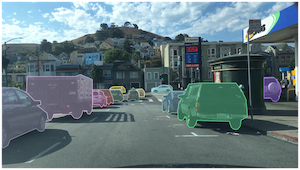} 
\end{tabular}
\egroup

\caption{\small Overview of our dataset. Our dataset includes a diverse set of driving videos under various weather conditions, time, and scene types. The dataset also comes with a rich set of annotations: scene tagging, object bounding box, lane marking, drivable area, full-frame semantic and instance segmentation, multiple object tracking, and multiple object tracking with segmentation.}
\vspace{-5mm}
\label{fig:teaser}
\end{figure*}

For autonomous driving applications, however, leveraging the power of deep learning is not as simple due to the lack of comprehensive datasets. Existing datasets for autonomous driving~\cite{kitti,cityscapes,vistas} are limited in one or more significant aspects, including the scene variation, the richness of annotations, and the geographic distribution. Additionally, models trained on existing datasets tend to overfit specific domain characteristics~\cite{rebuffi2017learning}.

Real-world applications require performing a combination of perception tasks with different complexities, instead of only \emph{homogeneous} multiple tasks with the same prediction structure~\cite{visualdec,Zamir_2018_CVPR,rob,mccann2018natural}. Although it may be feasible to label a large number of images with simple annotations such as drivable areas and object bounding boxes~\cite{pascal,mscoco}, it remains challenging to obtain more complicated annotations such as instance segmentation~\cite{acuna2018efficient}, not to mention multi-object detection and segmentation tracking~\cite{mots,mot16}. As a result, even though a considerable amount of effort has been put into constructing large-scale visual datasets, research on those complicated tasks is still limited to small datasets~\cite{cityscapes,kitti}. In production environments, it is also unclear how to allocate resources for various annotations to support the applications requiring \emph{heterogeneous} tasks with various output structures.

We aim to facilitate algorithmic study on large-scale diverse visual data and multiple tasks. We build BDD100K, a new, diverse, and large-scale dataset of visual driving scenes, together with various tasks, to overcome the limitations. We have been able to collect and annotate the largest available dataset of annotated driving scenes, consisting of over 100K diverse video clips. BDD100K covers more realistic driving scenarios and captures more of the ``long-tail'' of appearance variation and pose configuration of categories of interest in diverse environmental domains. Our benchmarks are comprised of ten tasks: image tagging, lane detection, drivable area segmentation, road object detection, semantic segmentation, instance segmentation, multi-object detection tracking, multi-object segmentation tracking, domain adaptation, and imitation learning, as shown in Figure~\ref{fig:teaser}. These diverse tasks make the study of heterogeneous multitask learning possible. In our benchmarks, the models can perform a series of tasks with increasing complexities.

We conduct extensive evaluations of existing algorithms on our new benchmarks. Special attention is paid to multitask learning in homogeneous, cascaded, and heterogeneous settings. Our experiments present many new findings, made possible by the diverse set of tasks on a single dataset. Our benchmark models on heterogeneous multitask learning shed light on the challenges of designing one single model to support multiple tasks.

The major contributions of our paper are: 1) a comprehensive diverse 100K driving video dataset supporting tasks of multiple complexities, which can serve as an evaluation benchmark for computer vision research for autonomous driving; 2) a benchmark for heterogeneous multitask learning and baseline studies to facilitate future study.
\vspace{-2mm}

\section{Related Works}

\minisection{Visual datasets} are necessary for numerous recognition tasks in computer vision. Especially with the advent of deep learning methods, large scale visual datasets, such as~\cite{imagenet,lsun,places,vistas}, are essential for learning high-level image representations. They are general-purpose and include millions of images with image-level categorical labels. These large datasets are useful in learning representations for image recognition, but most of the complex visual understanding tasks in the real world require more fine-grained recognition such as object localization and segmentation~\cite{pascal}. Our proposed dataset provides these multi-granularity annotations for more in-depth visual reasoning. In addition, we provide these annotations in the context of videos, which provides an additional dimension of visual information. Although large video datasets exist~\cite{activitynet,youtube,ucf101}, they usually are restricted to image-level labels.

\minisection{Driving datasets} have received increasing attention in the recent years, due to the popularity of autonomous vehicle technology. The goal is to understand the challenge of computer vision systems in the context of self-driving. Some of the datasets focus on particular objects such as pedestrians~\cite{dollar2009pedestrian,zhang2017citypersons}. Cityscapes~\cite{cityscapes} provides instance-level semantic segmentation on sampled frames of videos collected by their own vehicle. RobotCar~\cite{robotcar} and KITTI~\cite{kitti} also provide data of multiple sources such as LiDAR scanned points. Because it is very difficult to collect data that covers a broad range of time and location, the data diversity of these datasets is limited. For a vehicle perception system to be robust, it needs to learn from a variety of road conditions in numerous cities. Our data was collected from the same original source as the videos in~\cite{xu2017end}. However, the primary contribution of our paper is the video annotations with benchmarks on heterogeneous tasks. Mapillary Vistas~\cite{vistas} provides fine-grained annotations for user uploaded data, which is much more diverse with respect to location. However, these images are one-off frames that are not placed in the context of videos with  temporal structure. Like Vistas, our data is crowdsourced, however, our dataset is collected solely from drivers, with each annotated image corresponding to a video sequence, which enables interesting applications for modeling temporal dynamics.

\begin{figure*}[tp]
  \centering
  \includegraphics[width=0.9\linewidth]{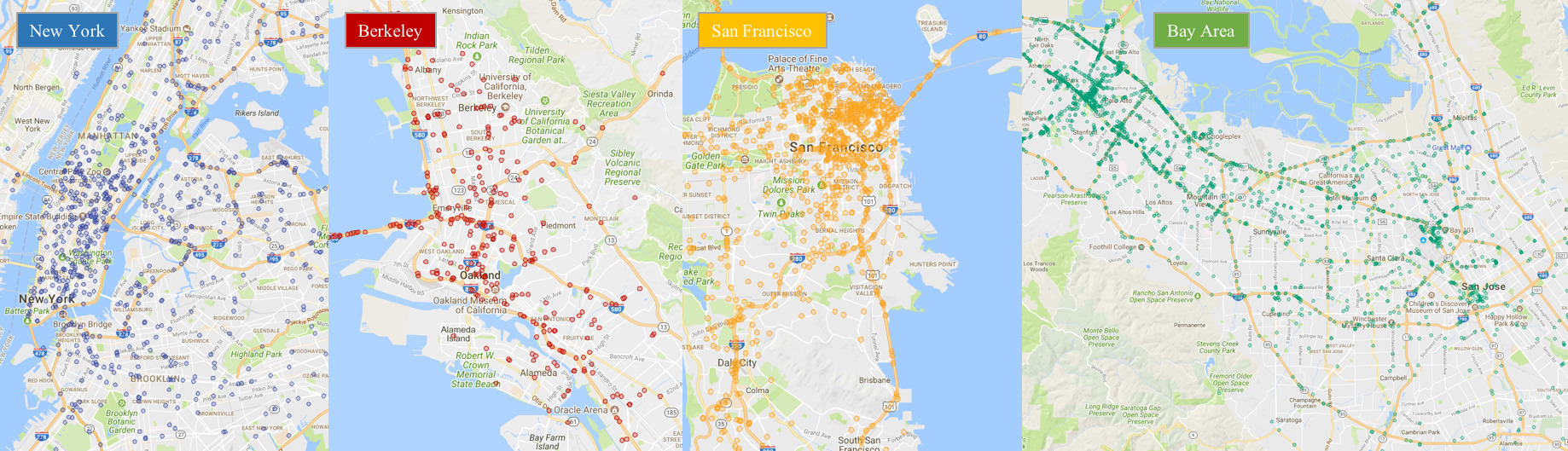}
  \caption{\small Geographical distribution of our data sources. Each dot represents the starting location of every video clip. Our videos are from many cities and regions in the populous areas in the US. \vspace{-3mm}}
  \label{fig:geo}
\end{figure*}

\begin{figure*}[tp]
        \centering
        \begin{subfigure}[b]{0.5\textwidth}
                \includegraphics[width=\linewidth]{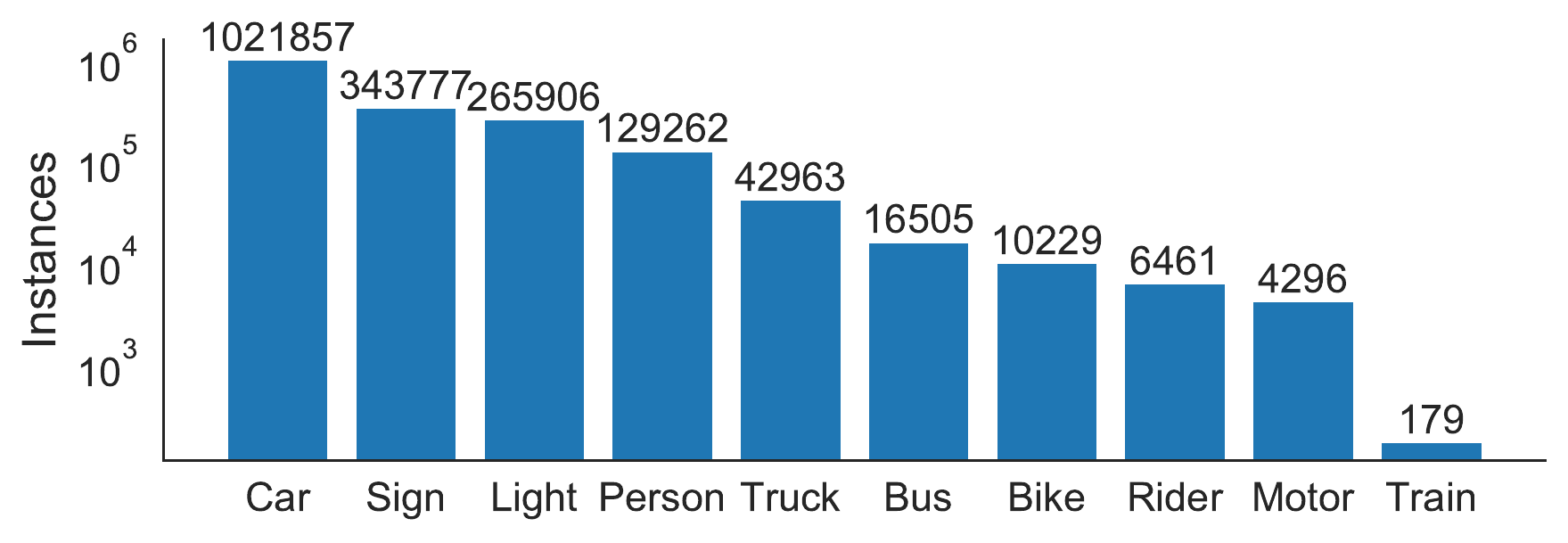}
                \vspace{-7mm}
                \caption{Number of instances in each category.}
                \label{fig:bbox_instance}
        \end{subfigure}%
        \begin{subfigure}[b]{0.2\textwidth}
                \includegraphics[width=\linewidth]{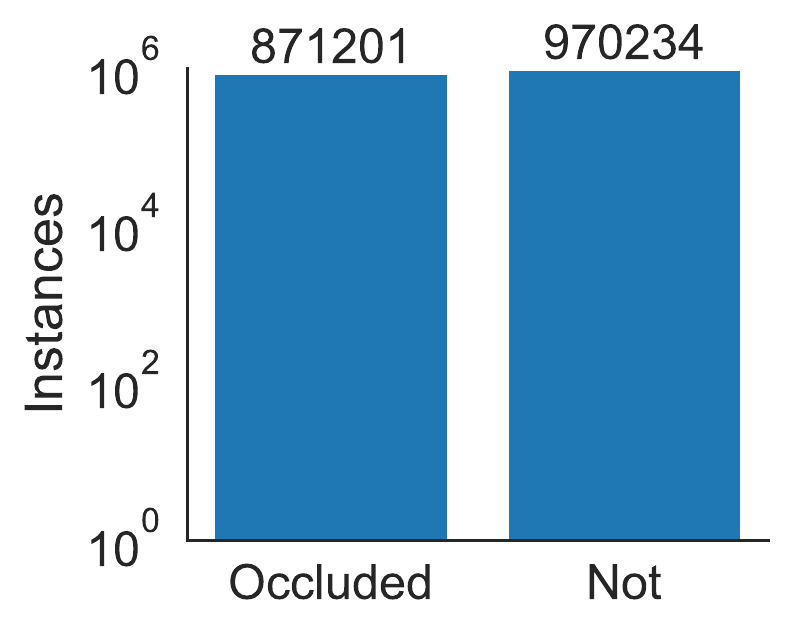}
                 \vspace{-7mm}
                \caption{Occlusion}
                \label{fig:occlusion}
        \end{subfigure}%
        \begin{subfigure}[b]{0.2\textwidth}
            \includegraphics[width=\linewidth]{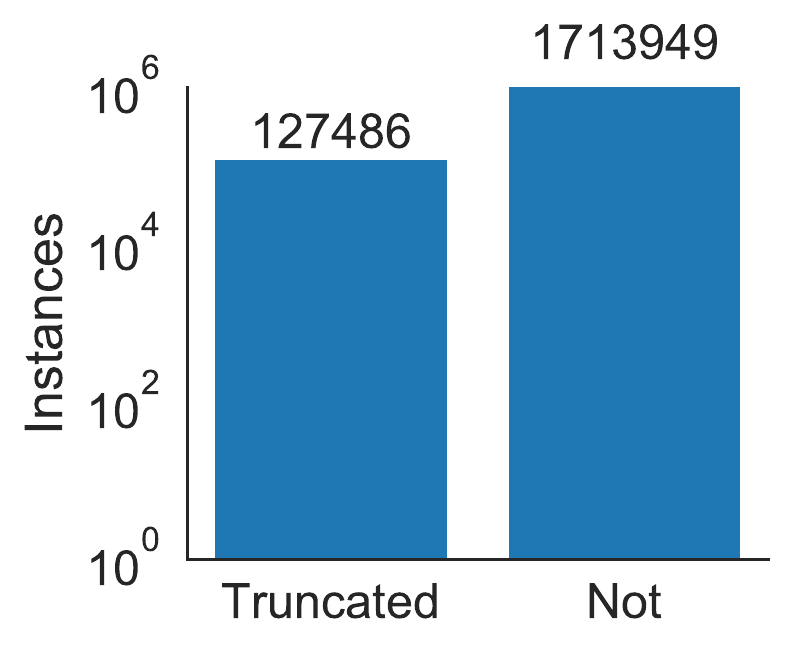}
              \vspace{-7mm}
                \caption{Truncation}
                \label{fig:truncation}
        \end{subfigure}%
        \vspace{-2mm}
        \caption{\small Instance statistics of our object categories. (a) Number of instances of each category, which follows a long-tail distribution. (b) Roughly half of the instances are occluded. (c) About 7$\%$ of the instances are truncated.
        \vspace{-4mm}}\label{fig:image_seg}
\end{figure*}

\minisection{Multitask Learning} aims to improve generalization of a certain task by learning from other tasks~\cite{caruana1997multitask, mitchell1980need}. 
It has been widely studied in machine learning~\cite{caruana1997multitask, evgeniou2004regularized}. 
The growing interests in learning the relationship between tasks gives rise to a number of multitask and transfer learning training benchmarks and challenges. Robust Vision Challenge~\cite{rob} features six vision challenges, where a single model is expected to produce results on multiple vision tasks. Zamir~\etal~\cite{Zamir_2018_CVPR} investigate the dependency structure among twenty-six visual tasks by transfer learning. McCann~\etal~\cite{mccann2018natural} present a challenge with ten natural language processing tasks, and proposes a model that solves all by formulating each task as question answering. Similar to McCann~\etal~\cite{mccann2018natural}, existing multitask and transfer learning setups are homogeneous in output structures. The tasks can be formulated as pixel-level or low-dimensional classification and regression. BDD100K contains multiple tasks including pixel-level, region-based, and temporally aware tasks, opening the door for heterogeneous multitask learning.
\vspace{-2mm}

\section{BDD100K}

\vspace{-2mm}
We aim to provide a large-scale diverse driving video dataset with comprehensive annotations that can expose the challenges of street-scene understanding.
To achieve good diversity, we obtain our videos in a crowd-sourcing manner uploaded by tens of thousands of drivers, supported by Nexar~\footnote{https://www.getnexar.com}. The dataset contains not only images with high resolution (720p) and high frame rate (30fps), but also GPS/IMU recordings to preserve the driving trajectories. In total, we have 100K driving videos (40 seconds each) collected from more than 50K rides, covering New York, San Francisco Bay Area, and other regions as shown in Figure~\ref{fig:geo}. 

 The dataset contains diverse scene types such as city streets, residential areas, and highways. Furthermore, the videos were recorded in diverse weather conditions at different times of the day. The videos are split into training (70K), validation (10K) and testing (20K) sets. The frame at the 10\textsuperscript{th} second in each video is annotated for image tasks and the entire sequences are used for tracking tasks.

\vspace{-1mm}
\subsection{Image Tagging}
\begin{figure}[h]
    \centering
    \includegraphics[width=\linewidth]{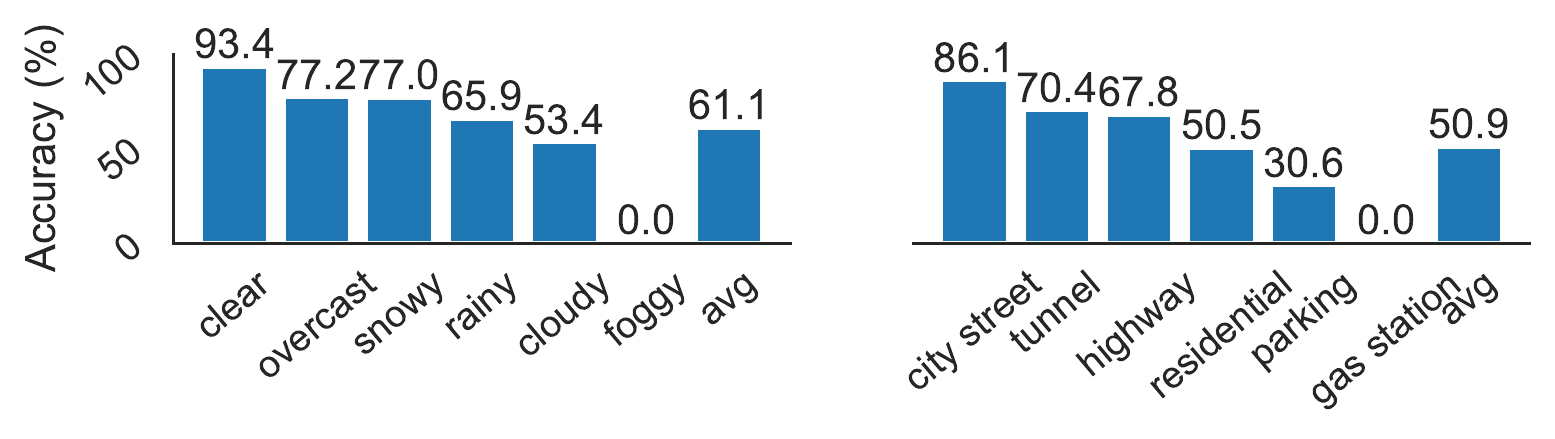}
    \vspace{-5mm}
    \caption{\small Image tagging classification results using DLA-34.\vspace{-4mm}}
    \label{fig:tagging}
\end{figure}

\vspace{-1mm}
We have collected image-level annotation on six weather conditions, six scene types, and three distinct times of day, for each image. The videos contain large portions of extreme weather conditions, such as snow and rain. They also include a diverse number of different scenes across the world. Notably, our dataset contains approximately an equal number of day-time and night-time videos. Such diversity allows us to study domain transfer and generalize our object detection model well on new test sets. Detailed distributions of images with weather, scene, and day hours tags are shown in the Appendix. We provide image tagging classification results using DLA-34~\cite{dla} in Figure~\ref{fig:tagging}. The average classification accuracy across different weather and scenes are around 50 to 60$\%$.

\begin{figure*}[tp]
    \centering
    \adjustbox{width=0.7\linewidth}{
    \begin{tabular}{c@{\hspace{1pt}}c@{\hspace{1pt}}c}
        \includegraphics[width=0.3\linewidth]{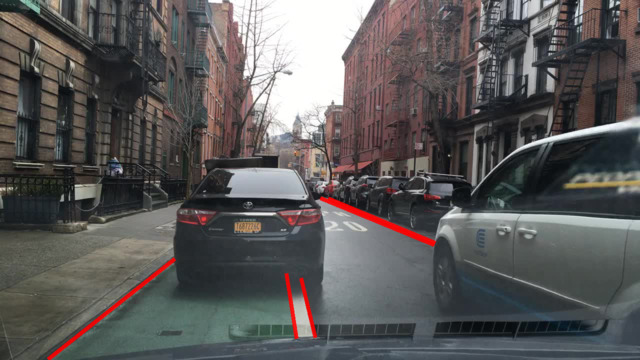} & 
        \includegraphics[width=0.3\linewidth]{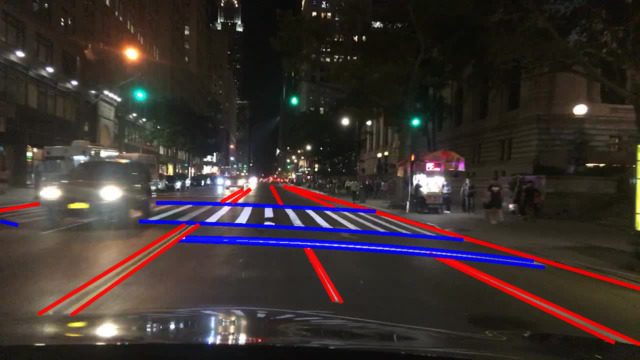} & 
        \includegraphics[width=0.3\linewidth]{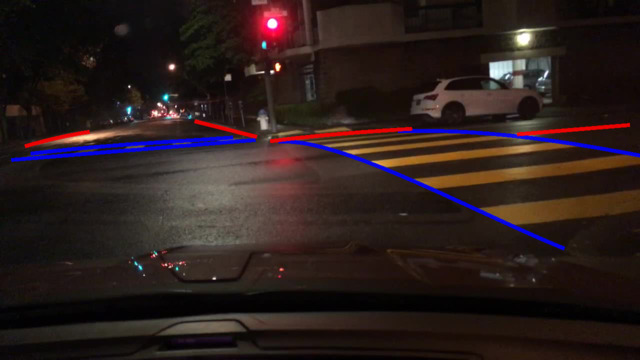} \\
    \end{tabular}
    }
    \caption{\small Examples of lane marking annotations. Red lanes are vertical and blue lanes are parallel. Left: we label all the visible lane boundaries. Middle: not all marking edges are lanes for vehicles to follow, such as pedestrian crossing. Right: parallel lanes can also be along the current driving direction.\vspace{-2mm}}
    \label{fig:lane_examples}
\end{figure*}

\begin{figure*}[tp]
    \centering
    \adjustbox{width=.9\linewidth}{
    \begin{tabular}{c@{\hspace{1pt}}c@{\hspace{1pt}}c@{\hspace{1pt}}c}
        \includegraphics[width=0.3\linewidth]{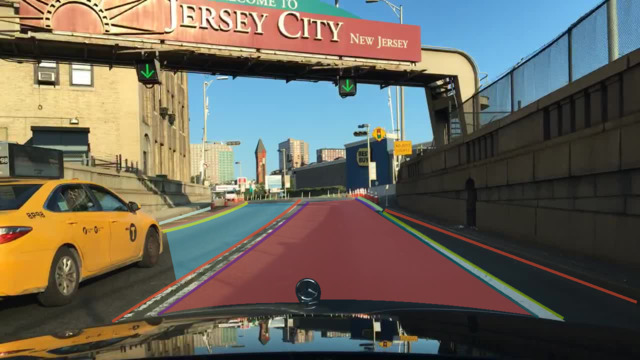} & 
        \includegraphics[width=0.3\linewidth]{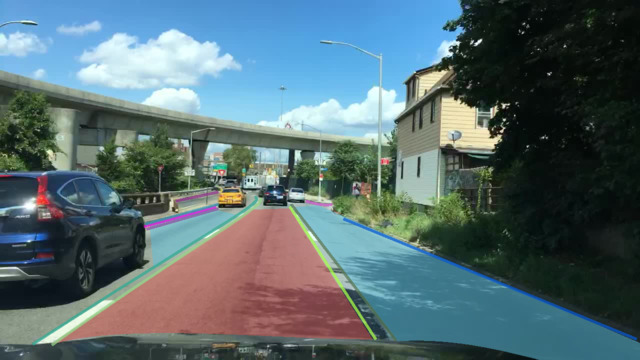} &
        \includegraphics[width=0.3\linewidth]{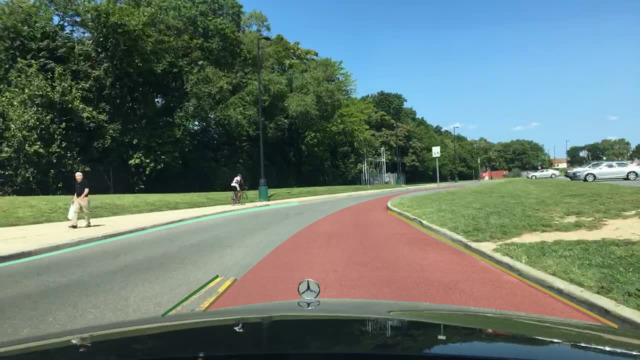} &
        \includegraphics[width=0.3\linewidth]{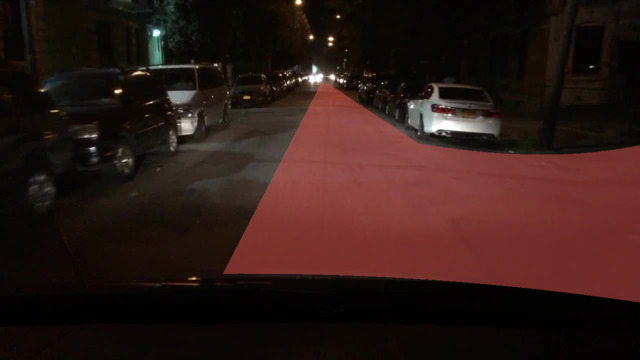} \\
    \end{tabular}
    }
    \caption{\small Examples of drivable areas. Red regions are directly drivable and the blue ones are alternative. Although drivable areas can be confined within lane markings, they are also related to locations of other vehicles shown in the right two columns.\vspace{-4mm}}
    \label{fig:drivable_examples}
\end{figure*}

\subsection{Object Detection}
Locating objects is a fundamental task for not only autonomous driving but the general visual recognition.
We provide bounding box annotations of 10 categories for each of the reference frames of 100K videos. The instance statistics is shown in Figure~\ref{fig:bbox_instance}. We provide visibility
attributes including ``occluded'' and ``truncated'' in Figure~\ref{fig:occlusion} and
Figure~\ref{fig:truncation}. 

\subsection{Lane Marking}

The lane marking detection is critical for vision-based vehicle localization and trajectory planning. However,
available datasets are often limited in scale and diversity. For example, the Caltech Lanes
Dataset~\cite{Aly2008Real} only contains 1,224 images, and the Road Marking Dataset~\cite{Wu2012A} has 1,443
images labeled in 11 classes of lane markings. The most recent work, VPGNet~\cite{lee2017vpgnet}, consists of
about 20,000 images taken during three weeks of driving in Seoul. 

Our lane markings (Figure~\ref{fig:lane_examples}) are labeled with 8 main categories: road curb,
crosswalk, double white, double yellow, double other color, single white, single yellow, single other color. The \textit{other} categories are ignored during evaluation.
We label the attributes of continuity (full or dashed) and direction (parallel or perpendicular). Shown
in Table~\ref{tab:lane_label_comp}, our lane marking annotations cover a diverse set of classes. Detailed
distributions of types of lane markings and drivable areas are shown in the Appendix.
\vspace{-2mm}
\begin{table}[htp]
    \centering
    \adjustbox{width=.8\linewidth}{
    \begin{tabular}{l@{\hspace{2pt}}|@{\hspace{2pt}}c@{\hspace{2pt}}c@{\hspace{2pt}}c}
    \toprule
    Datasets & Training & Total & Sequences\\ \midrule
    Caltech Lanes Dataset~\cite{Aly2008Real} & - & 1,224 & 4 \\
        Road Marking Dataset~\cite{Wu2012A} & - & 1,443 & 29 \\
         KITTI-ROAD~\cite{fritsch2013new} & 289 & 579 & - \\
         VPGNet~\cite{lee2017vpgnet} & 14,783 & 21,097 & - \\
         \midrule
         BDD100K & 70,000 & 100,000 & 100,000\\
         \bottomrule
    \end{tabular}}
    \caption{\small Lane marking statistics. Our lane marking annotations are significantly richer and are more diverse.\vspace{-2mm}}
    \label{tab:lane_label_comp}
\end{table}

\subsection{Drivable Area}

Lanes alone are not sufficient to decide road affordability for driving. Although most of the time, the vehicle should stay between the lanes, it is common that no clear lane marking exists. In addition, the road area is shared with all other vehicles, but a lane can not be driven on if occupied. All these conditions beyond lane markings direct our driving decisions and thus are relevant for designing autonomous driving algorithms.

Our drivable areas are divided into two different categories: \emph{directly drivable area} and \emph{alternatively drivable area}. The directly drivable area is what the driver is currently driving on -- it is also the region where the driver has priority over other cars or the right of the way. In contrast, alternatively drivable area is a lane the driver is currently not driving on, but able to do so via changing lanes. Although the directly and alternatively drivable areas are visually indistinguishable, they are functionally different, and require the algorithms to recognize blocking objects and scene context. Some examples are shown in Figure~\ref{fig:drivable_examples}. The distribution of drivable region annotations is shown in the Appendix. Not surprisingly, on highways or city streets, where traffic is closely regulated, drivable areas are mostly within lanes and they do not overlap with the vehicles or objects on the road. However, in residential areas, the lanes are sparse. Our annotators can find the drivable areas based on the surroundings.

\subsection{Semantic Instance Segmentation}

We provide fine-grained, pixel-level annotations for images from each of the 10,000 video clips randomly sampled from the whole dataset. Each pixel is given a label and a corresponding identifier denoting the instance number of that object label in the image. Since many classes (e.g., sky) are not amenable to being split into instances, only a small subset of class labels are assigned instance identifiers. The entire label set 
consists of 40 object classes, which are chosen to capture the diversity of objects in road scenes as well as maximizing the number of labeled pixels in each image. Besides a large number of labels, our dataset exceeds previous efforts in terms of scene diversity and complexity. The whole set is split into 3 parts: 7K images for training, 1K images for validation, and 2K images for testing. The distribution of classes in the semantic instance segmentation dataset is shown in the Appendix.

\subsection{Multiple Object Tracking}

To understand the temporal association of objects within the videos, we provide a multiple object tracking (MOT) dataset including 2,000 videos with about 400K frames. Each video is approximately 40 seconds and annotated at 5 fps, resulting in approximately 200 frames per video. We observe a total number of 130.6K track identities and 3.3M bounding boxes in the training and validation set. The dataset splits are 1400 videos for training, 200 videos for validation and 400 videos for testing. Table~\ref{tab:box_tracking_label_comp} shows the comparison of BDD100K with previous MOT datasets. Our tracking benchmark provides one order-of-magnitude bigger than the previously popular tracking dataset, MOT17~\cite{mot16}. A recent dataset released by Waymo~\cite{waymo_2019} has fewer tracking sequences (1150 vs 2000) and fewer frames (230K vs 398K) in total, compared to ours. But Waymo data has more 2D boxes (9.9M vs 4.2M), while ours has better diversity including different weather conditions and more locations. Distributions of tracks and bounding boxes by category are shown in the Appendix.

\begin{table}[htp]
\centering
\adjustbox{width=.78\linewidth}{
\small
\begin{tabular}{l@{\hspace{4pt}}|c@{\hspace{4pt}}c@{\hspace{4pt}}c@{\hspace{4pt}}c@{\hspace{4pt}}}
    \toprule
   Datasets          & Frames & Sequences & Identities & Boxes     \\ \midrule
KITTI~\cite{kitti}  & 8K       & 21            & 917           & 47K \\
MOT17~\cite{mot16} & 34K & 21            & 1,638      & 337K   \\
\midrule
BDD100K & 318K         & 1,600      &           131K    & 3.3M \\ 
\bottomrule
\end{tabular}}
\caption{\small MOT datasets statistics of training and validation sets. Our dataset has more sequences, frames, identities as well as more box annotations.\vspace{-2mm}}
    \label{tab:box_tracking_label_comp}
\end{table}

\begin{figure}[tp]
    \centering
    \vspace{4mm}
    \setlength{\tabcolsep}{1pt}
    \adjustbox{width=\linewidth}{
    \begin{tabular}{c@{\hspace{2pt}}c@{\hspace{2pt}}c}
        \includegraphics[height=0.35\linewidth]{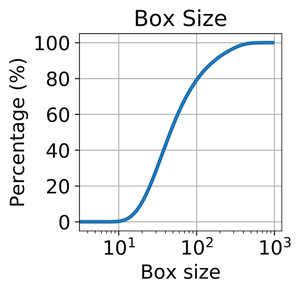} &
        \includegraphics[height=0.35\linewidth]{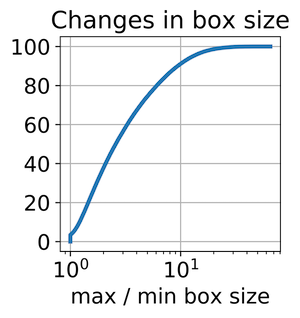} &
        \includegraphics[height=0.35\linewidth]{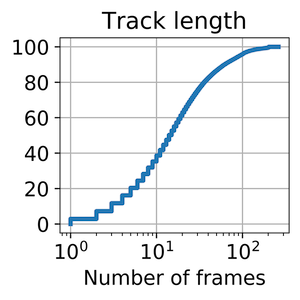} \\
    \end{tabular}
    }
    \caption{\small Cumulative distributions of the box size (left), the ratio between the max and min box size for each track (middle) and track length (right). Our dataset is more diverse in object scale.\vspace{-1mm}}
    \label{fig:mot-track}
\end{figure}

\begin{figure}[htp]
    \vspace{-2mm}

    \centering
    \setlength{\tabcolsep}{1pt}

    \begin{tabular}{cc}
        \includegraphics[width=0.45\linewidth]{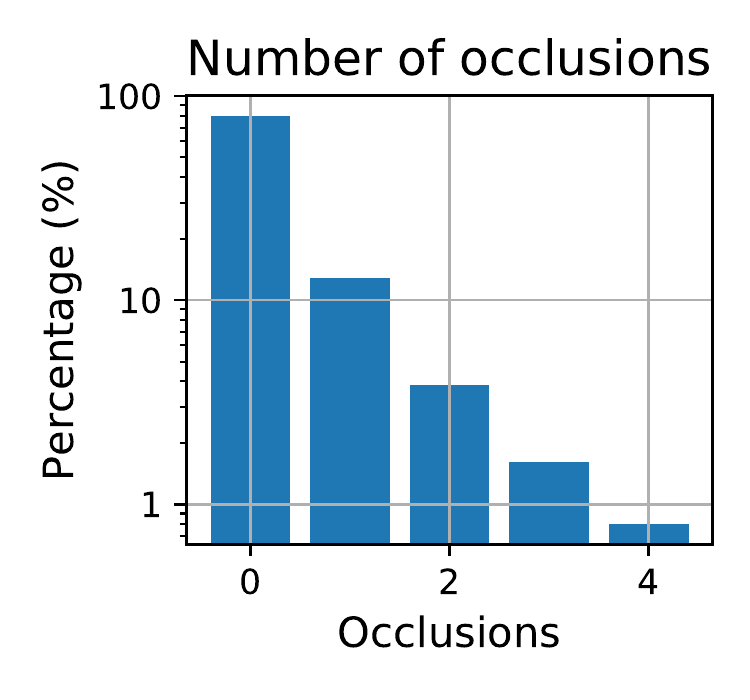} & \hspace{3mm}
        \includegraphics[width=0.45\linewidth]{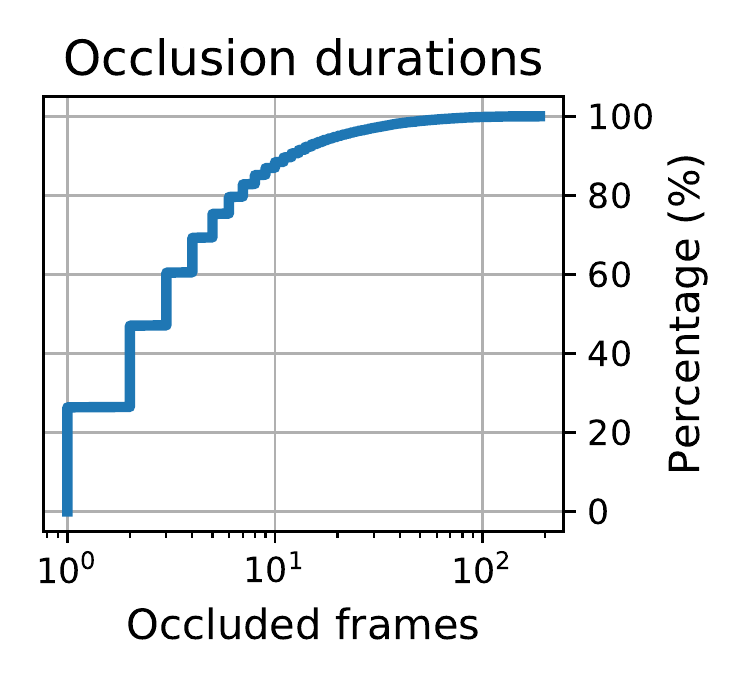} \\
    \end{tabular}
    \vspace{-2mm}
    \caption{\small Number of occlusions by track (left) and number of occluded frames for each occlusion (right).  Our dataset covers complicated occlusion and reappearing patterns. \vspace{-2mm}}
    \label{fig:mot-occlusion}
    
\end{figure}

BDD100K MOT is diverse in object scale. Figure~\ref{fig:mot-track} (left) plots the cumulative distribution of box size, defined as $\sqrt{wh}$ for a bounding box with width $w$ and height $h$. Figure~\ref{fig:mot-track} (middle) shows the cumulative distribution of the ratio between the maximum box size and the minimum box size along each track, and  Figure~\ref{fig:mot-track} (right) shows that of the length of each track. The distributions show that the MOT dataset is not only diverse in visual scale among and within tracks, but also in the temporal range of each track.

Objects in our tracking data also present complicated occlusion and reappearing patterns are shown in Figure~\ref{fig:mot-occlusion}. An object may be fully occluded or move out of the frame, and then reappear later. We observe 49,418 occurrences of occlusion in the dataset, or one occurrence of occlusion every 3.51 tracks. Our dataset shows the real challenges of object re-identification for tracking in autonomous driving.

\subsection{Multiple Object Tracking and Segmentation}

We further provide a multiple object tracking and segmentation (MOTS) dataset with 90 videos. We split the dataset into 60 training videos, 10 validation videos, and 20 testing videos.

\begin{table}[htp]
\centering
\adjustbox{width=0.9\linewidth}{
\begin{tabular}{l@{\hspace{4pt}}|c@{\hspace{4pt}}c@{\hspace{4pt}}c@{\hspace{4pt}}c@{\hspace{4pt}}c@{\hspace{4pt}}}
    \toprule
  Datasets   & Frames & Seq. & Identities & Ann. & Ann. / Fr.     \\ \midrule
KITTI MOTS~\cite{mots}  & 8K          & 21             & 749        & 38K & 4.78  \\
MOTS Challenge~\cite{mots} & 2.9K     & 4             & 228        & 27K & 9.40  \\
\midrule
DAVIS 2017~\cite{davis17} & 6.2K      & 90                & 197           &  - & -      \\
YouTube VOS~\cite{ytvos} & 120K      & 4.5K                & 7.8K           &  197K & 1.64      \\
\midrule
BDD100K MOTS & 14K     & 70              & 6.3K      & 129K & 9.20 \\
\bottomrule
\end{tabular}
}
\caption{\small Comparisons with other MOTS and VOS datasets. }
\label{tab:seg_tracking_label_comp}
\end{table}

\begin{figure*}[htp]
    \centering
    \setlength{\tabcolsep}{1pt}
    \begin{tabular}{cccc}        \includegraphics[width=0.24\textwidth]{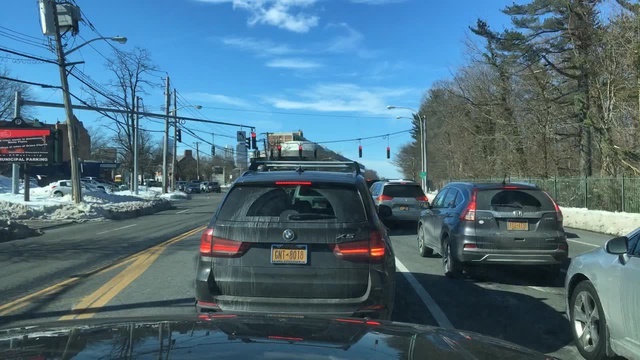} &
        \includegraphics[width=0.24\textwidth]{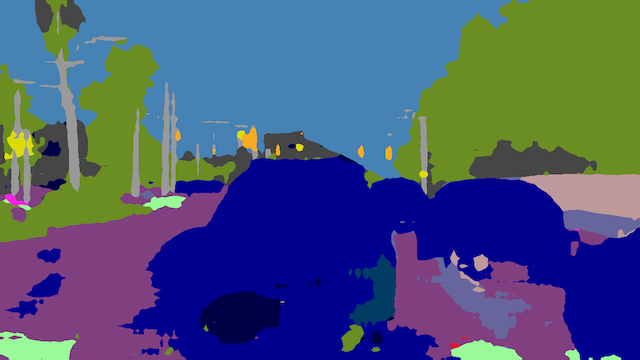} &
        \includegraphics[width=0.24\textwidth]{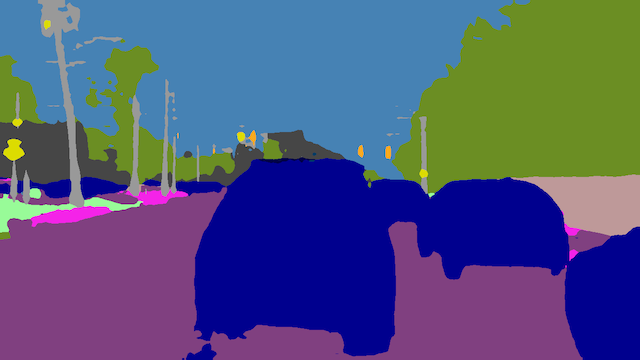} &
        \includegraphics[width=0.24\textwidth]{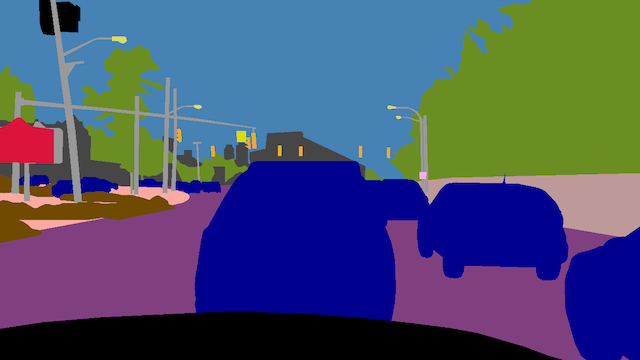} \\
        
        \includegraphics[width=0.24\textwidth]{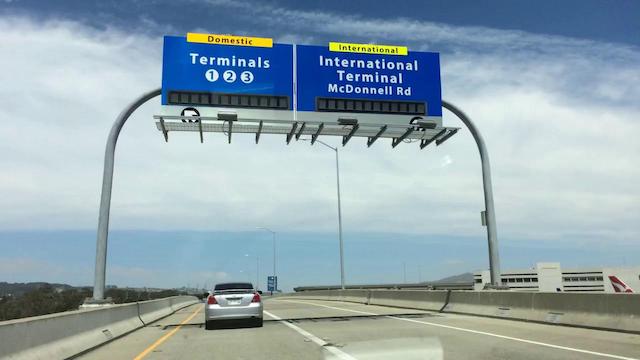} &
        \includegraphics[width=0.24\textwidth]{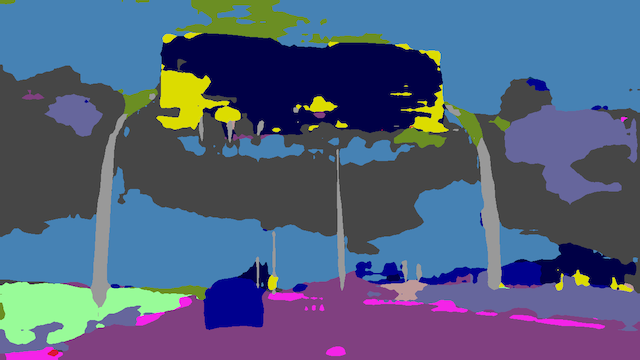} &
        \includegraphics[width=0.24\textwidth]{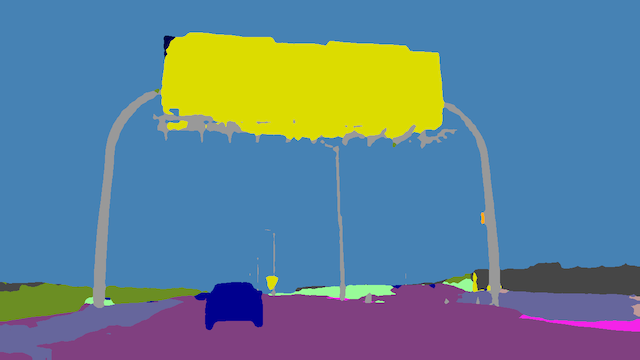} &
        \includegraphics[width=0.24\textwidth]{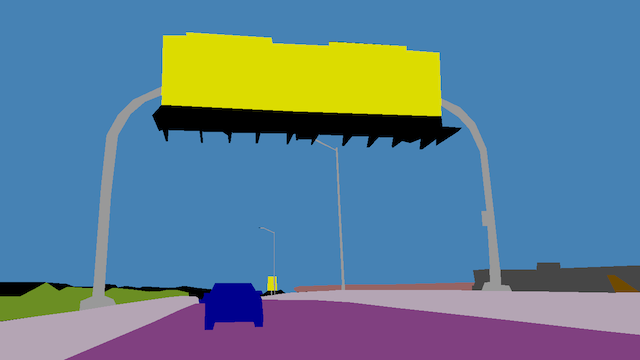} \\
        (a) \small{Our Image} & \small{(b) Trained on Cityscapes} & \small{(c) Trained on Ours} & \small{(d) Ground Truth} \\
    \end{tabular}
    \vspace{-2mm}
    \caption{\small Visual comparisons of the same model (DRN~\cite{drn}) trained on different datasets. We find that there is a dramatic domain shift between Cityscapes and our new dataset. For example, due to infrastructure difference, the model trained on Cityscapes is confused by some simple categories such as sky and traffic signs.}
    \label{fig:cityscapes_vis}
\end{figure*}

\begin{table*}[htb]
\centering
\adjustbox{width=.65\linewidth}{
\begin{tabular}{l|ccc|l|ccc}
\toprule
\backslashbox{Train}{Test}& City & Non-City & Val  &\backslashbox[30mm]{Train}{Test} & Daytime & Non-Daytime & Val \\ \midrule
City-30K & 29.5 & 26.5 & 28.8  & Daytime-30K & 30.6 & 23.6 & 28.1\\
Non-City-30K & 24.9 & 24.3 & 24.9 & Non-Daytime-30K & 25.9 & 25.3 & 25.6\\
Random-30K & 28.7 & 26.6 & 28.3 & Random-30K & 29.5 & 26.0 & 28.3\\
\bottomrule
\end{tabular}}
\caption{\small Domain discrepancy experiments with object detection. We take the images from one domain and report testing results in AP on the same domain or the opposite domain. We can observe significant domain discrepancies, especially between daytime and nighttime.\vspace{-5mm}}
\label{tab:detection_transfer}
\end{table*}

Table~\ref{tab:seg_tracking_label_comp} shows the details of the BDD MOTS dataset and the comparison with existing multiple object tracking and segmentation (MOTS) and video object segmentation (VOS) datasets. MOTS aims to perform segmentation and tracking of multiple objects in crowded scenes. Therefore, MOTS datasets like KITTI MOTS and MOTS Challenge~\cite{mots} require denser annotations per frame and therefore are smaller in size than VOS datasets. BDD100K MOTS provides a MOTS dataset that is larger than the KITTI and MOTS Challenge datasets, with the number of annotations comparable with the large-scale YouTube VOS \cite{ytvos} dataset. Detailed distributions of the MOTS dataset by category are shown in the Appendix.

\subsection{Imitation Learning}

GPS/IMU recordings in our dataset show the human driver action given the visual input and the driving trajectories. We can use those recordings as a demonstration supervision for the imitation learning algorithms and use perplexity to measure the similarity of driving behaviors on the validation and testing set. We refer to Xu~\etal~\cite{xu2017end} for details on the evaluation protocols. Visualizations of the driving trajectories are shown in the Appendix.

\section{Diversity}
One distinct feature of our data is diversity, besides video and scale. We can study new challenges that the diversity brings to existing algorithms and how our data complements existing datasets. We conduct two sets of experiments on object detection and semantic segmentation. In object detection experiments, we study the different domains within our dataset. While in semantic segmentation, we investigate the domains between our data and Cityscapes~\cite{cityscapes}.

\subsection{Object Detection}
\label{sec:detection-domain}

Our dataset has an advantage in diversity, compared to other popular driving datasets. We investigate the influence of domain differences on object detection. The whole dataset is partitioned into several domains based on time of day and scene types. City street and daytime are chosen as validation domains. The training sets have the same number of images (30K) in the training set. We then train Faster-RCNN~\cite{ren2015faster} based on ResNet-50 on those domains and evaluate the result with COCO API \cite{mscoco}.

We find that there is indeed a domain discrepancy between image sets from different conditions, as shown in Table~\ref{tab:detection_transfer}. The difference between city and non-city is significant, but the gap between daytime and nighttime is much bigger. Although this is not completely surprising, the results indicate that more work is necessary to bridge the gap.

\vspace{-1mm}
\subsection{Semantic Segmentation}
\vspace{-1mm}
We also compare the models trained on Cityscapes and ours, to understand the difference between our new datasets and existing driving datasets. Cityscapes data is collected in German cities, while our data is mainly from the US.
We observe that there is a dramatic domain shift between the two datasets for semantic segmentation models. The models perform much worse when tested on a different dataset. 
This suggests that even for the domain of other datasets, our new dataset is complementary, which augments existing datasets. Figure~\ref{fig:cityscapes_vis} shows the discrepancy visually. We can observe that the model trained on Cityscape can not recognize the traffic sign in the US.

\section{Multitask Learning}
\vspace{-1mm}
BDD100K gives the opportunity to study joint solution for the heterogeneous tasks. In this section, we investigate the effects of modeling various tasks jointly with the same base model. We study how to utilize diversity and quantity of simple labels to improve the accuracy of the complicated tasks, such as from object detection to tracking.

\subsection{Homogeneous Multitask Learning}
\label{sec:homogeneous}
We first investigate the effects of jointly performing tasks with similar output structures. The BDD100K lane marking and drivable area datasets share the same set of 70K training images. Drivable area annotations consist of 2 foreground classes and lane marking annotations have 3 attributes (direction, continuity, and category). We formulate the detection of drivable area as segmentation and lane marking as contour detection. We evaluate drivable area segmentation by mean IoU, and lane marking by the Optimal Dataset Scale F-measure (ODS-F) for each category of the three attributes using the Structured Edge Detection Toolbox~\cite{DollarICCV13edges} with tolerance $\tau =$ 1, 2, and 10 pixels. We employ morphological thinning for each score threshold during evaluation.

We employ DLA-34~\cite{dla} as the base model for the segmentation tasks. We implement the segmentation head with four $3\times3$ convolution blocks followed by an $1\times1$ convolution to produce segmentation maps in a 4x down-sampled scale, and use bilinear interpolation to upsample the output to the original scale. For lane marking, we use three segmentation heads for the three attributes. We employ the weighted cross-entropy loss with foreground weight 10 for the lane marking heads, and the gradient-based non-maximum suppression for post-processing. We construct three train sets with 10K, 20K and the full 70K images and report the evaluation results of models trained on 
individual tasks and both tasks in Table~\ref{tab:lane_drivable}. Full evaluation results for lane marking are shown in the Appendix.

\begin{table}[htp]
\centering
\adjustbox{width=\linewidth}{
\begin{tabular}{l@{\hspace{4pt}}|c@{\hspace{4pt}}c@{\hspace{4pt}}c@{\hspace{4pt}}c@{\hspace{4pt}}|c@{\hspace{4pt}}c@{\hspace{4pt}}c@{\hspace{4pt}}}
    \toprule
Training Set  &   \multicolumn{4}{c|}{Lane ODS-F ($\tau=10$) (\%)}     & \multicolumn{3}{c}{Drivable IoU (\%)} \\
                  & dir. \hspace{1mm} & cont. \hspace{1mm} & cat. \hspace{1mm} & mean  & direct & altern. & mean\\ \midrule
Lane 10K          & 49.29         & 47.85          & 39.08        & 45.41  & -            & -             & -                     \\
Drive 10K      & -            & -             & -           & -        & 73.10            & 55.36             & 64.23            \\
Lane+Drive 10K & 53.97        & 52.59         & 44.65       & 50.40   & 74.69            & 54.06             & 64.37           \\ \midrule
Lane 20K           & 57.36              & 55.85               & 49.88             & 54.36          & -            & -             & -             \\
Drive 20K       & -            & -             & -           & -      & 79.00            & 63.27             & 71.13               \\
Lane+Drive 20K & 57.19         & 55.64        & 49.50        & 54.11    & 79.39            & 64.06             & 71.73                \\ \midrule
Lane 70K & 57.50 & 55.87 & 50.08 & 54.48 & -            & -             & -          \\
Drive 70K      & -            & -             & -           & -       & 79.40            & 63.33             & 71.37             \\
Lane+Drive 70K & 57.35        & 55.76         & 49.63       & 54.24    & 79.72            & 64.70             & 72.21             \\ 
\bottomrule
\end{tabular}}
    \caption{\small Evaluation results of homogeneous multitask learning on lane marking and drivable area segmentation. We train lane marking, drivable area segmentation and the joint training of both on training splits with 10K, 20K, and the full 70K images.}
    \vspace{-3mm}
    \label{tab:lane_drivable}
\end{table}

We observe that when training with only 10K images, the mean ODS-F score of lane marking prediction improves from 45.41 to 50.40 when jointly training with the drivable area task. However, the improvement of jointly training on the drivable area detection task, from 64.23 to 64.37, is marginal compared to the individual task.  
As we increase the number of training images to 20K and 70K, the difference between
jointly training and single-task training becomes insignificant, though the performance numbers are
generally higher than those trained on 10K images. 

One hypothesis for the results is that the drivable area detection task and the 
lane marking task share a similar prediction structure, referred as the homogeneous tasks, and therefore the additional supervision may fail to bring new information to each individual task. These results further motivate us to study multitask learning of heterogeneous tasks with diverse prediction structure and annotation types in this work.

\subsection{Cascaded Multitask Learning}

Certain tasks such as object tracking and instance segmentation are more time-consuming to annotate. But they can depend on predictions of simple tasks. This connection has been studied as cascaded multitask learning. For example, more accurate object detection can locate the object candidates better for tracking. A natural question is whether to spend all the annotation efforts for the complicated tasks, or to allocate some resources for the basic tasks.
\begin{table}[htp]
\centering
\small
\adjustbox{max width=\linewidth}{
\begin{tabular}{l@{\hspace{4pt}}|@{\hspace{4pt}}c@{\hspace{4pt}}c@{\hspace{4pt}}c@{\hspace{4pt}}c@{\hspace{4pt}}c@{\hspace{4pt}}}
    \toprule
Training Set  & AP & AP$_{50}$ & AP$_{75}$\\ \midrule
Inst-Seg      &       21.8       &        40.5       &      20.5       \\
Inst-Seg + Det    &      \textbf{24.5}       &     \textbf{45.4}        &      \textbf{21.6}      \\
\bottomrule
\end{tabular}}
    \caption{\small Evaluation results for instance segmentation when joint training with the object detection set. Additional localization supervision can improve instance segmentation significantly.\vspace{-4mm}}
    \label{tab:ins_seg_results}
\end{table} 

\begin{table}[htp]
\centering
\adjustbox{max width=\linewidth}{
\small
\begin{tabular}{l@{\hspace{4pt}}|@{\hspace{4pt}}c@{\hspace{4pt}}c@{\hspace{4pt}}c@{\hspace{4pt}}c@{\hspace{4pt}}c@{\hspace{4pt}}c@{\hspace{4pt}}}
    \toprule
Training Set  & AP & MOTA   & MOTP & IDS\\ \midrule
MOT & 28.1 & 55.0 & 84.0 & \textbf{8386} \\
MOT + Det           & \textbf{30.7} & \textbf{56.7} & \textbf{84.1} & 9098  \\
\bottomrule
\end{tabular}
}
    \caption{\small Evaluation results for multiple object tracking cascaded with object detection. AP is the detection metric. Even though the tracking set has much more boxes, the model can still benefit from the diverse instance examples in the detection set. \vspace{-2mm}}
    
    \label{tab:box_tracking_results}
\end{table}

\begin{table*}[htbp]
\begin{center}

\ra{1.2}
\adjustbox{max width=\linewidth}{
\small
  \begin{tabular}{@{}l@{\hspace{5mm}}*{21}{@{\hspace{2mm}}l}@{\hspace{5mm}}c@{\hspace{2mm}} c}
  Training Set &  \ver{Road} & \ver{Sidewalk} & \ver{Building} & \ver{Wall} &
    \ver{Fence} & \ver{Pole} & \ver{Light} & \ver{Sign} & \ver{Vegetation} & \ver{Terrain} & \ver{Sky} & \ver{Person} & \ver{Rider}
    & \ver{Car} & \ver{Truck} & \ver{Bus} & \ver{Train} &
    \ver{Motorcycle} & \ver{Bicycle} & \ver{mean IoU} \\
  \hline

Sem-Seg & 94.3 & 63.0 & 84.9 & \textbf{25.7} & \textbf{45.8} & \textbf{52.6} & 56.2 & 54.1 & \textbf{86.4} & 45.1 & 95.3 & 62.4 & 22.1 & 90.2 & 50.5 & 68.3 & 0 & 35.5 & \textbf{49.9} & 56.9 \\
Sem-Seg + Det & 94.3 & 62.5 & \textbf{85.2} & 24.5 & 41.1 & 51.5 & \textbf{63.1} & \textbf{57.9} & 86.2 & \textbf{47.4} & \textbf{95.5} & \textbf{64.6} & \textbf{28.1} & \textbf{90.8} & \textbf{52.9} & \textbf{70.7} & 0 & \textbf{43.4} & 48.9 & \textbf{58.3} \\
Sem-Seg + Lane + Driv &  \textbf{94.8} & \textbf{65.8} & 84.1 & 22.6 & 40.2 & 49.3 & 51.9 & 49.7 & 85.8 & 46.2 & 95.3 & 60.8 & 7.1 & 89.9 & 47.8 & 66.9 & 0 & 27.5 & 27.5 & 53.3\\
\hline
  \end{tabular}
}
\end{center}

\vspace{-3mm}
\caption{\small Evaluation results for semantic segmentation. We explore segmentation joint-training with different tasks. Detection can improve the overall accuracy of segmentation, although their output structures are different. However, although Lane and Drivable area improve the segmentation of road and sidewalk, the overall accuracy drops. \vspace{-3mm}}
  \label{tab:sem_seg_results}

\end{table*}

\minisection{Object detection and instance segmentation}. The BDD instance segmentation dataset contains 7K images, whereas the detection dataset has 70K images. We first study whether adding more object detection annotations can help instance segmentation. We use Mask R-CNN~\cite{he2017mask} with ResNet-50~\cite{he2016deep} as the backbone, and train detection and instance segmentation in a batch-level round-robin manner. As shown in Table~\ref{tab:ins_seg_results}, AP increases from 21.8 to 24.5 with joint training. The instance segmentation model is able to learn better object appearance features and localization from the detection set with a much richer diversity of images and object examples. Zhou~\etal~\cite{zhou2020} explore the shape priors in the detection supervision and improve the semi-supervised instance segmentation results further.

\minisection{MOT and object detection.} BDD100K MOT has 278K training frames from 1,400 videos, whereas the detection set contains 70K images sampled from 70K videos. For the detection and MOT models, we use a modified version of Faster R-CNN~\cite{ren2015faster} with a shared DLA-34~\cite{dla} backbone. The implementation details of the tracking model are shown in the Appendix.
Table~\ref{tab:box_tracking_results} shows that joint training of detection and multiple object tracking improves the single-task MOT model with detection AP increasing from 28.1 to 30.7 and MOTA from 55.0 to 56.7, with a slight increase in identity switch. 

\minisection{Semantic segmentation with other tasks.} Following a similar manner, we fine-tune a base semantic segmentation model by jointly training semantic segmentation with detection and lane marking/drivable area as shown in Table~\ref{tab:sem_seg_results}. We observe that training with the additional 70K object detection dataset improves the overall mIoU from 56.9 to 58.3, with the improvement mostly attributed to the object classes that are present in the object detection dataset.  When jointly training with the lane marking and drivable area sets, the IOU of the stuff classes (e.g., road and sidewalk) improves though the overall IOU across all classes decreases. 

To summarize, adding more annotations to the simple tasks in the task cascade can help improve the performances of the complicated tasks that require more expensive labels. 

\subsection{Heterogeneous Multitask Learning}

The ultimate goal of our benchmark is to study how to perform all the heterogeneous tasks together for autonomous driving. To understand the potential and difficulty, we study joint training for multiple object tracking and segmentation, a downstream task to object detection, instance segmentation, and multiple object tracking. Since the MOTS dataset requires time-consuming instance segmentation annotations at each frame, the dataset is relatively limited in video diversity, with 12K frames from 60 videos in the training set. We aim to improve the performance on the task of MOTS by leveraging the diversity from the detection set with 70K images from 70K videos, the MOT set with 278K frames from 1,400 videos, and the instance segmentation set with 7K images from 7K videos.

We report instance segmentation AP and multi-object tracking and segmentation accuracy (MOTSA), precision (MOTSP), and other metrics used by \cite{mots} in Table \ref{tab:seg_tracking_results}. We first fine-tune the MOTS model from pre-trained models of upstream tasks. Compared with training MOTS from scratch, fine-tuning from the pre-trained instance segmentation model improves segmentation AP and MOTSP. Fine-tuning from the pre-trained MOT model, on the other hand, reduces identity switch (IDSW). The extra training examples from the instance segmentation and MOT datasets improve the segmentation and box propagation respectively, thus improving the overall MOTSA results by a large margin. We finally fine-tune the jointly trained detection and tracking model mentioned in Table \ref{tab:box_tracking_results} by jointly training the four tasks together. We achieve an overall segmentation AP of 23.3 and MOTSA of 41.4.

\begin{table}[htp]
\small
\centering
\vspace{1mm}
\adjustbox{max width=\linewidth}{
\begin{tabular}{p{2.5cm}@{\hspace{4pt}}|c@{\hspace{4pt}}@{\hspace{4pt}}c@{\hspace{4pt}}c@{\hspace{4pt}}c@{\hspace{4pt}}c@{\hspace{4pt}}c@{\hspace{4pt}}}
    \toprule
Training Set  & AP &  MOTSA   & MOTSP & FN & FP & IDSW \\ \midrule
MOTS (S)           & 13.0 & 30.4 & 81.8 & 8352 & \textbf{5116} & 566 \\
InstSeg (I) + MOTS & 18.7 & 33.7 & \textbf{81.9}  & 6810 & 5611 & 965 \\
MOT (T) + MOTS & 19.7 & 40.3 & 79.8 & 5698 & 5967 & \textbf{390}  \\ \midrule
Det + T + I + S& \textbf{23.3} & \textbf{41.4} & {81.6} & {\textbf{5132}} & {6228} & {472} \\
\bottomrule
\end{tabular}
}
\caption{\small MOTS evaluation results. Both instance segmentation AP and MOTS evaluation metrics are reported. Instance segmentation tracking is very hard to label, but we are able to use object detection, tracking, and instance segmentation to improve segmentation tracking accuracy significantly.\vspace{-5mm}}
    
    \label{tab:seg_tracking_results}
\end{table}

\section{Conclusion}
In this work, we presented BDD100K, a large-scale driving video dataset with extensive annotations for heterogeneous tasks.
We built a benchmark for heterogeneous multitask learning where the tasks have various prediction structures and serve different aspects of a complete driving system.
Our experiments provided extensive analysis to different multitask learning scenarios: homogeneous multitask learning and cascaded multitask learning. The results presented interesting findings about allocating the annotation budgets in multitask learning. We hope our work can foster future studies on heterogeneous multitask learning and shed light on this important direction.

{\small
\bibliographystyle{abbrv}
\bibliography{reference}
}

\section*{Appendix}
\appendix
\section{Dataset Details}
We present more details about our dataset and annotations in this section.

\subsection{Image Tagging}

\begin{figure*}[htp]
\adjustbox{width=\linewidth}{
        \vspace{-2mm}
        \begin{subfigure}[b]{0.375\textwidth}
                \includegraphics[width=\linewidth]{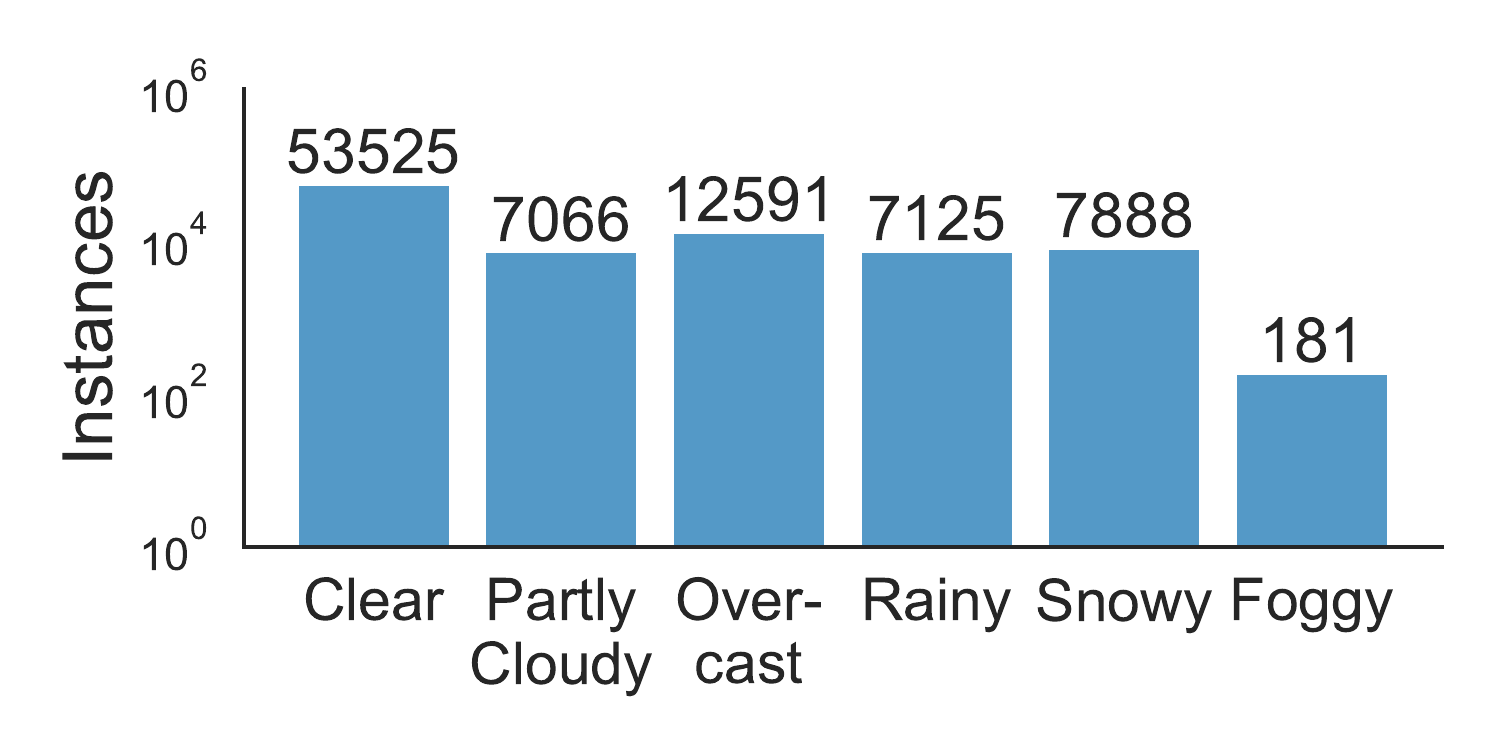}
                \vspace{-4mm}
                \caption{Weather}
                \label{fig:weather}
        \end{subfigure}%
        \begin{subfigure}[b]{0.375\textwidth}
                \includegraphics[width=\linewidth]{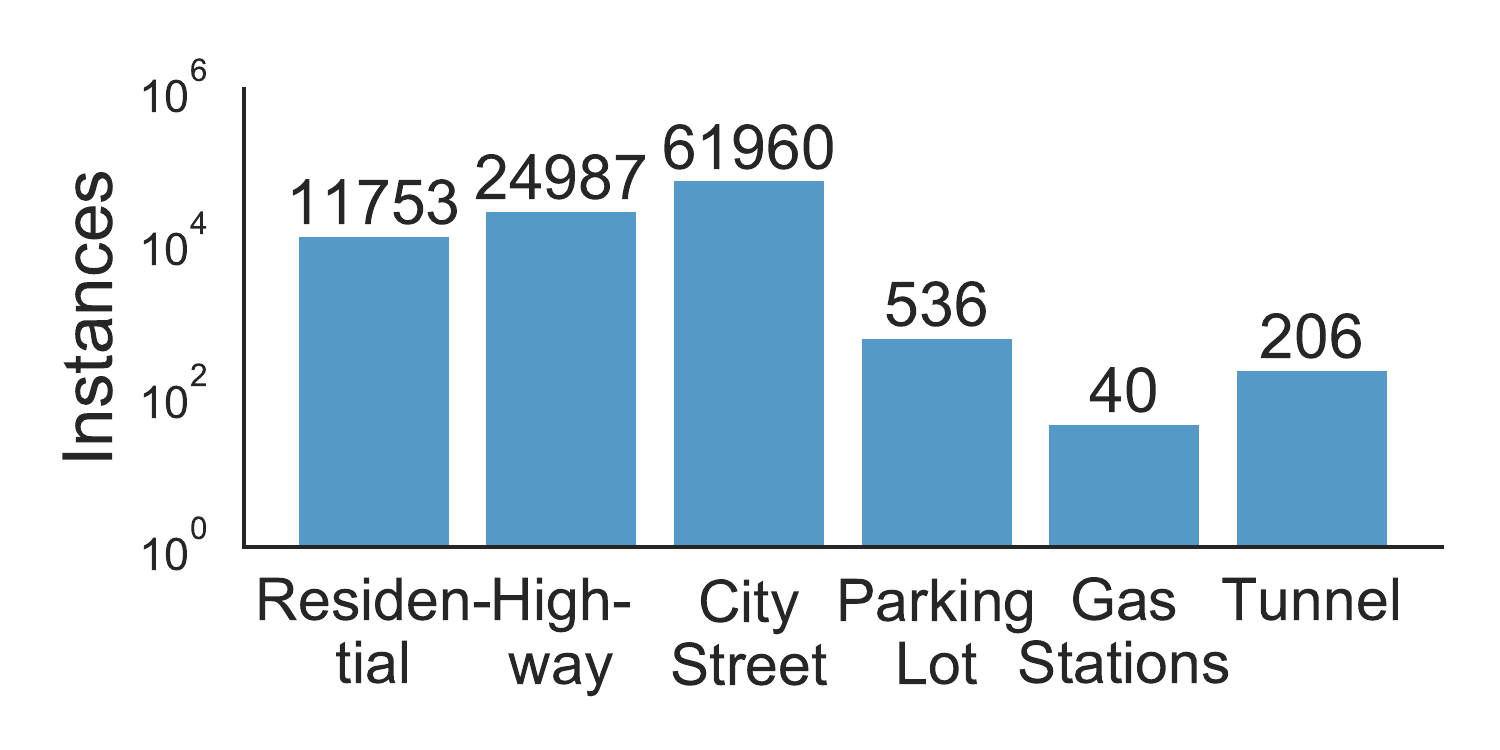}
                 \vspace{-4mm}
                \caption{Scene}
                \label{fig:scene}
        \end{subfigure}%
        \begin{subfigure}[b]{0.25\textwidth}
                \includegraphics[width=\linewidth]{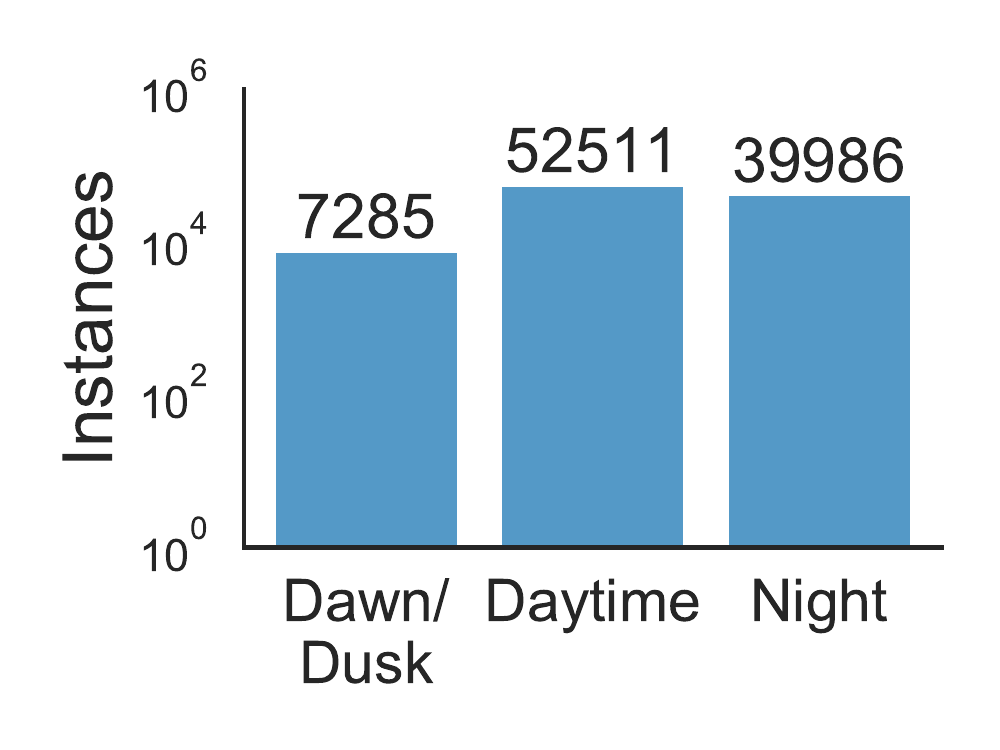}
              \vspace{-6mm}
                \caption{Hours}
                \label{fig:hours}
        \end{subfigure}%
}
        \vspace{-4mm}
        \caption{\small Distribution of images in weather, scene, and day hours categories.         \vspace{-3mm}
        }
        \label{fig:image_tagging}
\end{figure*}

Figure~\ref{fig:image_tagging} shows the distribution of weather, scene, and time of day attributes in BDD100K. The distribution demonstrates visual diversity of the images and thus provides an opportunity to study visual transfer between different domains.

\subsection{Object Detection}
\vspace{-2mm}
\begin{table}[htp]

    \centering
\adjustbox{width=.75\linewidth}{
    \begin{tabular}{c@{\hspace{4pt}}|@{\hspace{4pt}}c@{\hspace{4pt}}c@{\hspace{4pt}}c@{\hspace{4pt}}c}
    \toprule
         & Caltech~\cite{dollar2009pedestrian} & KITTI~\cite{kitti} & City~\cite{zhang2017citypersons} & Ours \\ \midrule
        \# persons & 1,273 & 6,336 & 19,654 & 86,047\\
        \# per image & 1.4 & 0.8 & 7.0 & 1.2 \\
    \bottomrule
    \end{tabular}}
     \caption{\small Comparisons on number of pedestrians with other datasets. The statistics are based on the training set in each dataset. \vspace{-2mm}}
    \label{tab:person_stats}
\end{table}

Table~\ref{tab:person_stats} compares the number of pedestrians of the detection dataset with other datasets. Our dataset has more examples of pedestrians, but because our dataset contains non-city scenes such as highways, the number of person per image is lower than Cityscapes.

\subsection{Lane Marking and Drivable Area}

Our choice of annotated lane attributes is based on their influence on driving decisions. The continuity of a lane marking is essential for making a ``driving-across'' decision, so we labeled it independently as an important attribute. Similarly, the direction of a lane marking is also significant for autonomous driving. For example, if a lane marking is parallel to the passing car, it may serve to guide cars and separate lanes; if it is perpendicular, it can be treated as a sign of deceleration or stop. The distribution of the number of annotations in varied driving scenes are shown in Figure~\ref{fig:lane1}, Figure~\ref{fig:lane2}, and Figure~\ref{fig:lane3}. The detailed evaluation results for the lane marking benchmark are in Table~\ref{tab:lane_eval}.

Drivable area detection is a new task, so we show results of a baseline method on the task here. First, the drivable area detection is converted to 3-way segmentation task (background, directly, and alternatively drivable) by ignoring the region ID. Then, we train DRN-D-22 model~\cite{drn} on the 70,000 training images. We find that after learning from the large-scale image dataset, the model learns to split the road according to the lanes and extrapolate the drivable area to unmarked space. The mIoU for directly and alternatively drivable areas is 79.4\% and 63.3\%. However, the same model can achieve much higher accuracy on road segmentation, which indicates that techniques beyond segmentation may be required to solve the drivable area problem.
\vspace{-3mm}

\begin{table}[htp]
    \centering
    \adjustbox{width=.9\linewidth}{
\begin{tabular}{l@{\hspace{2mm}}r@{\hspace{2mm}}*{10}{@{\hspace{2mm}}r}@{\hspace{2mm}}}
  & Total & \ver{person} & \ver{rider} & \ver{car} & \ver{truck} & \ver{bus} & \ver{train} & \ver{motorcycle} & \ver{bicycle} \\ \midrule
Tracks &  131K & 24K & 1.0K & 97K & 4.6K & 1.5K & 34 & 685 & 1.8K \\
Boxes  & 3.3M & 440K & 23K & 2.6M & 177K & 67K & 1.9K & 13K & 33K \\
Truncated &  346K & 18K & 1.4K & 284K & 26K & 11K & 557 & 1.4K & 3.1K \\
Occluded &  2.2M & 253K & 18K & 1.7M & 132K & 51K & 1.9K & 9.1K & 26K \\
\midrule
\end{tabular}}
\caption{\small Annotations of the BDD100K MOT dataset by category.\vspace{-2mm}}
\label{tab:mot-stats}
\end{table}

\begin{figure*}[htp]
    \centering
    \begin{tabular}{c@{\hspace{1pt}}c@{\hspace{1pt}}c}
        \includegraphics[width=0.3\linewidth]{images/drivable_seg/b205eb4d-f84aaa1a.jpg} & 
        \includegraphics[width=0.3\linewidth]{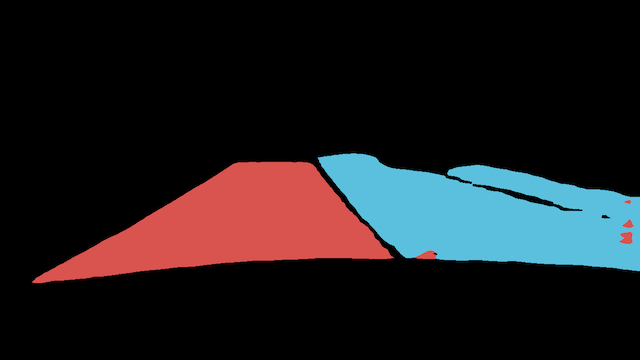} &
        \includegraphics[width=0.3\linewidth]{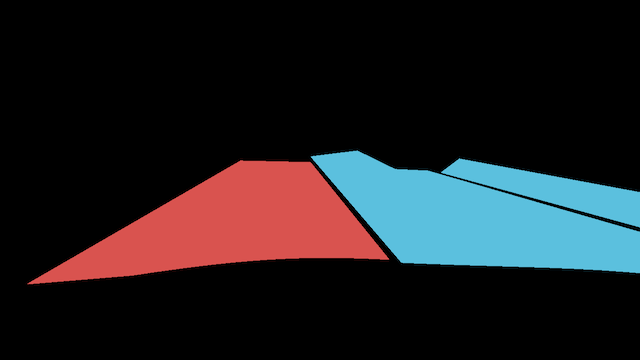} \\
        \includegraphics[width=0.3\linewidth]{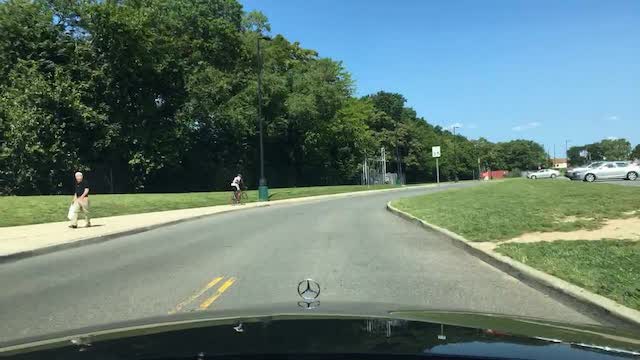} &
        \includegraphics[width=0.3\linewidth]{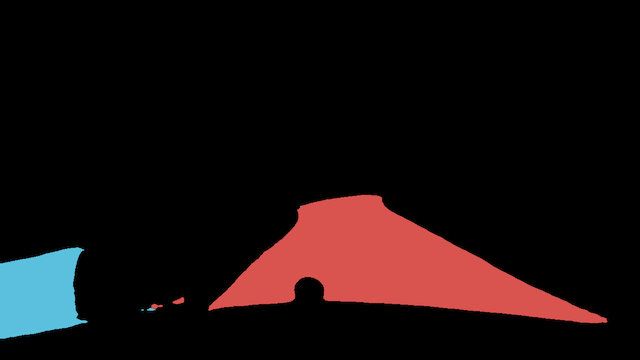} &
        \includegraphics[width=0.3\linewidth]{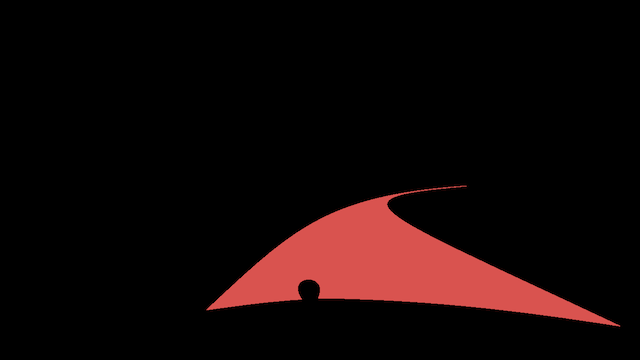} \\
        (a) Image & (b) DRN & (c) Ground truth 
    \end{tabular}
    \caption{\small Drivable area prediction by segmentation. The segmentation predicts the drivable area with lanes well, as shown in the top row. Also, we find that the segmentation model learns to interpolate in areas that has no lane markings.\vspace{-3mm}}
    
    \label{fig:drivable_seg}
\end{figure*}

\begin{figure*}[htp]
        \begin{subfigure}[b]{0.40\textwidth}
                \includegraphics[width=\linewidth]{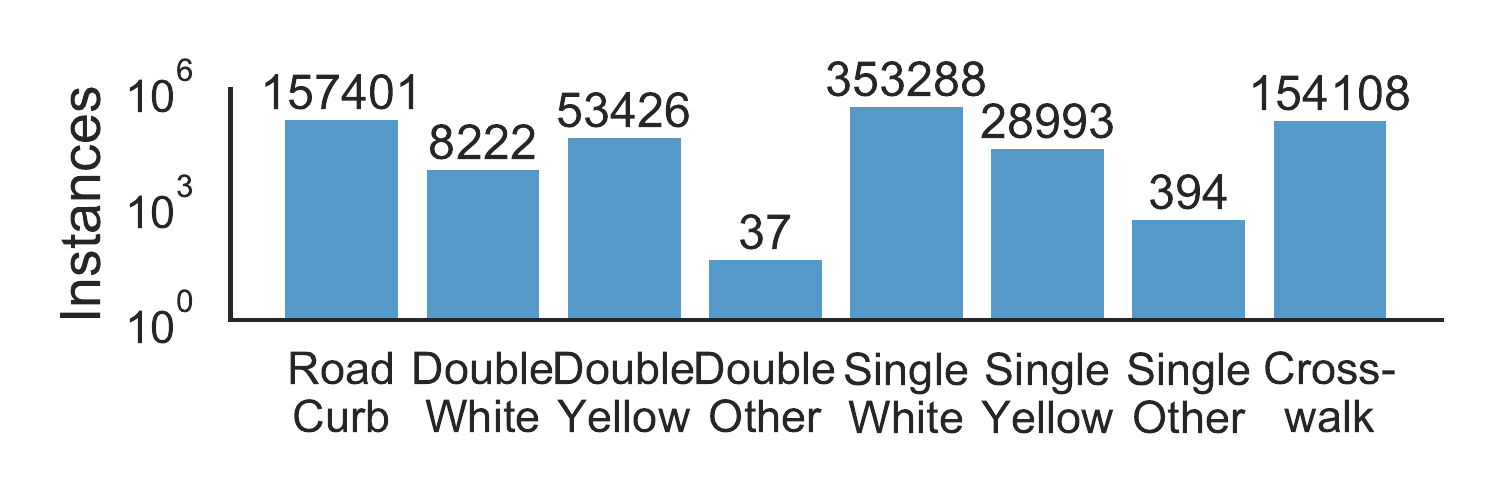}
                \vspace{-7mm}
                \caption{Lane category}
                \label{fig:lane1}
        \end{subfigure}%
        \begin{subfigure}[b]{0.20\textwidth}
                \includegraphics[width=\linewidth]{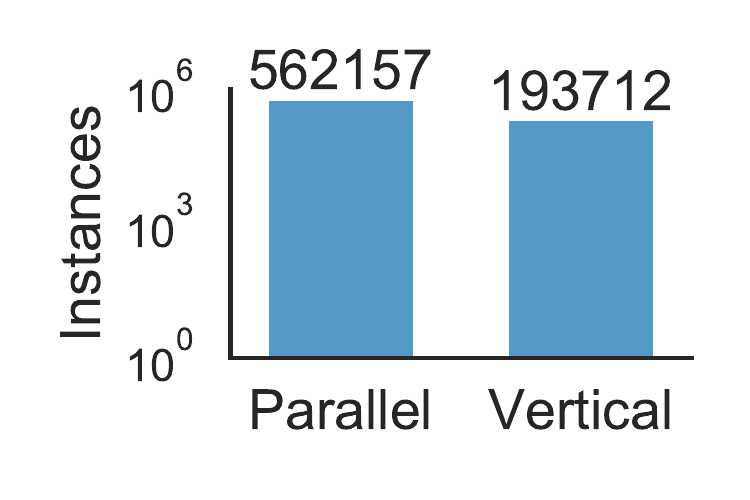}
                 \vspace{-7mm}
                \caption{Lane direction}
                \label{fig:lane2}
        \end{subfigure}%
        \begin{subfigure}[b]{0.20\textwidth}
                \includegraphics[width=\linewidth]{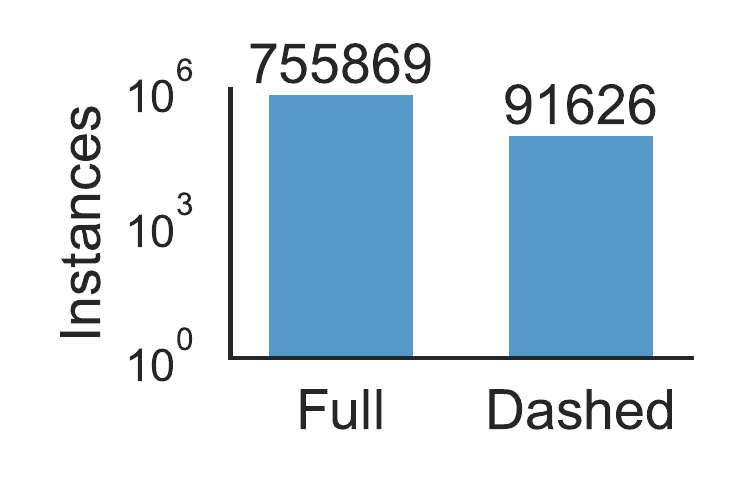}
              \vspace{-7mm}
                \caption{Lane continuity}
                \label{fig:lane3}
        \end{subfigure}%
        \begin{subfigure}[b]{0.20\textwidth}
                \includegraphics[width=\linewidth]{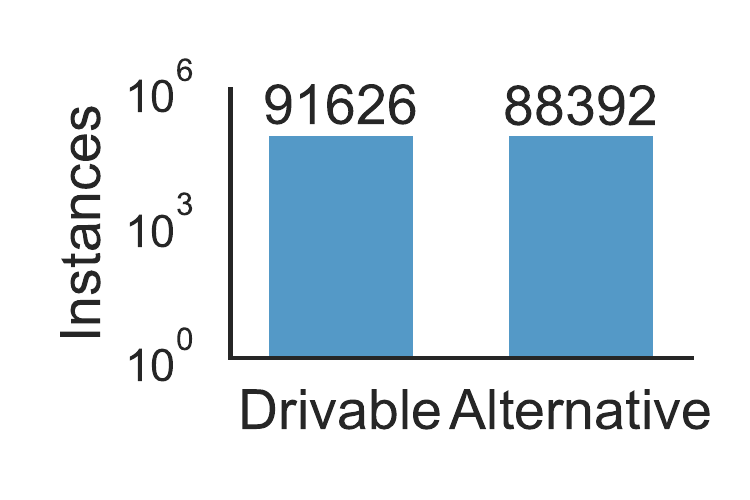}
              \vspace{-7mm}
                \caption{Drivable areas}
                \label{fig:drivable}
        \end{subfigure}
        \vspace{-7mm}
        \caption{\small Distribution of different types of lane markings and drivable areas.}\label{fig:all}
\end{figure*}

\subsection{GPS Trajectory}

\begin{figure*}
    \centering
    \adjustbox{width=0.9\linewidth}{
    \begin{tabular}{c@{\hspace{2pt}}c@{\hspace{2pt}}c@{\hspace{2pt}}c}
        \includegraphics[trim={1mm 1mm 0 0},clip,width=0.22\linewidth]{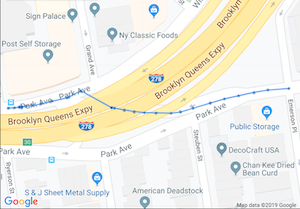} & 
        \includegraphics[trim={1mm 1mm 0 0},clip,width=0.22\linewidth]{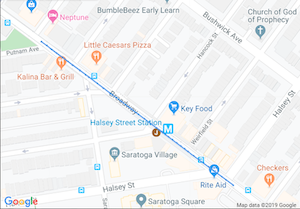} &
        \includegraphics[trim={1mm 1mm 0 0},clip,width=0.22\linewidth]{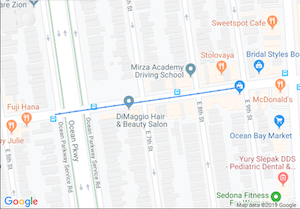}
        \includegraphics[trim={1mm 1mm 0 0},clip,width=0.22\linewidth]{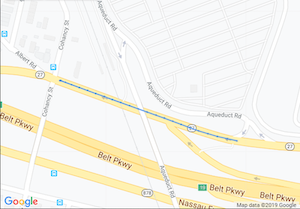}
    \end{tabular}
    }
    \caption{\small Trajectories of example driving videos.}
    \label{fig:gps_traj}
\end{figure*}

Figure~\ref{fig:gps_traj} shows GPS trajectories of example sequences. Our data presents diverse driving behaviors, like starting, stopping, turning and passing. The data is suitable to train and test imitation learning algorithms on real driving data.

\begin{figure*}[htp]
  \centering
  \vspace{-4mm}
  \includegraphics[width=.9\linewidth]{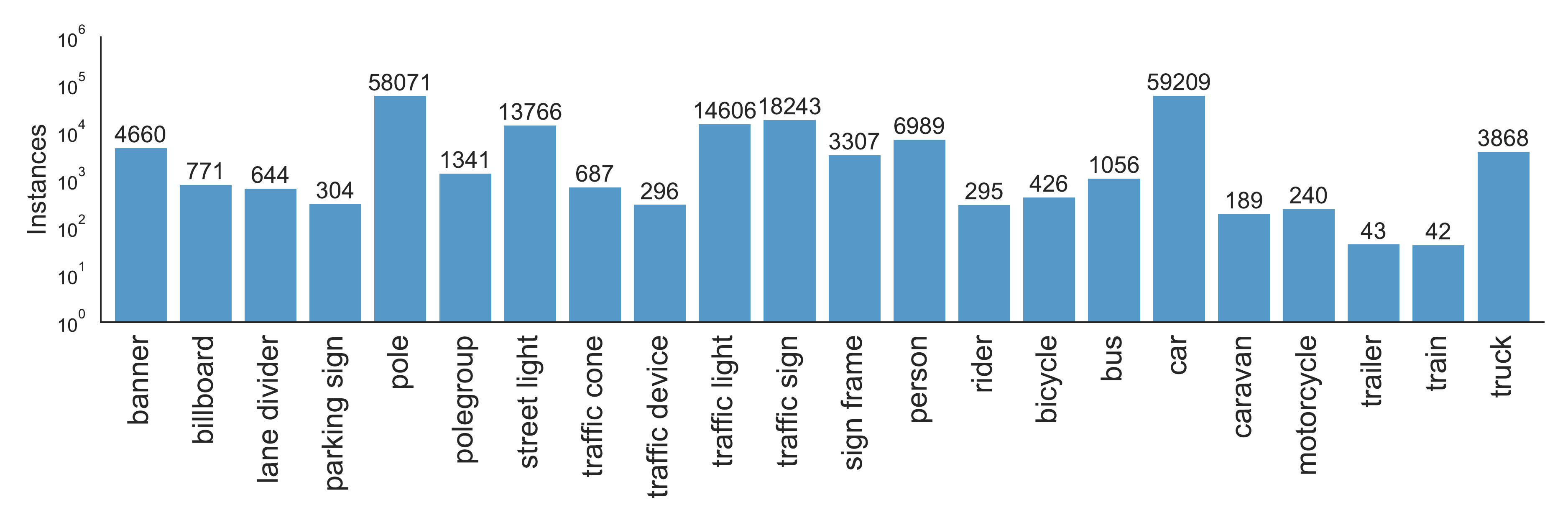}
  \vspace{-5mm}
  \caption{\small Distribution of classes in semantic instance segmentation. It presents a long-tail effect with more than 10 cars and poles per image, but only tens of trains in the whole dataset.\vspace{0mm}}
  \label{fig:seg_instance}
\end{figure*}

\subsection{Semantic Instance Segmentation}

Figure~\ref{fig:seg_instance} shows the distribution of number of instances observed in the segmentation annotations. BDD100K has a good coverage on rare categories (e.g. trailer, train) and large number of instances of common traffic objects such as persons and cars. We also observe long-tail effects on our dataset. There are almost 60 thousand car instances, a few hundred rider and motorcycle instances, and mere dozens of trailer and train instances.

Figure 9 in the main paper shows some segmentation examples produced by DRN-D-38. They also reveal some interesting properties of various domains. Probably because of the infrastructure differences between Germany and the US, the models trained on Cityscapes confuse some big structures in an unreasonable way, such as segmenting the sky as building as shown in the third row of the figure. The model is also confused by the US highway traffic sign. However, the same model trained on our dataset does not suffer these problems. Also, the model of Cityscapes may over-fit the hood of the data collecting vehicle and produces erroneous segmentation for the lower part of the images.

\subsection{Multiple Object Tracking and Segmentation}

Table \ref{tab:mot-stats} and Table \ref{tab:mots-stats} shows the label distributions by categories. Our bounding box tracking annotations cover more than one hundred thousand instances with more than two million bounding boxes, and the segmentation tracking set contains more than six thousand instances with one hundred thousand polygons. We showed in the paper submission, our tracking annotation set is one of the largest out there, in addition to our advantage in multitask and diversity.

\begin{table}[htp]
    \centering
\adjustbox{width=.9\linewidth}{
\small
\begin{tabular}{l@{\hspace{2mm}}r@{\hspace{2mm}}*{10}{@{\hspace{2mm}}r}@{\hspace{2mm}}}
  & Total & \ver{person} & \ver{rider} & \ver{car} & \ver{truck} & \ver{bus} & \ver{train} & \ver{motorcycle} & \ver{bicycle} \\ \midrule
Tracks &  6.3K & 1.8K & 31 & 4.0K & 215 & 93 & 4 & 21 & 76\\
Masks  & 129K & 22K & 894 & 93K & 7.6K & 4.0K & 117 & 369 & 1.4K \\
Truncated & 15K & 833 & 45 & 12K & 1.3K & 743 & 8 & 49 & 70 \\
Occluded &  85K & 13K & 793 & 61K & 5.7K & 3.1K & 116 & 292 & 970 \\
\midrule
\end{tabular}
}
\caption{\small Annotations of BDD100K MOTS by category.\vspace{-3mm}}
\label{tab:mots-stats}
\end{table}

\begin{figure*}[htp]
    \vspace{-5mm}

    \centering
    \setlength{\tabcolsep}{1pt}

    \begin{tabular}{ccccc}
        \includegraphics[width=0.19\textwidth]{images/seg_track/1-1.png} &
        \includegraphics[width=0.19\textwidth]{images/seg_track/1-2.png} &
        \includegraphics[width=0.19\textwidth]{images/seg_track/1-3.png} &
        \includegraphics[width=0.19\textwidth]{images/seg_track/1-4.png} &
        \includegraphics[width=0.19\textwidth]{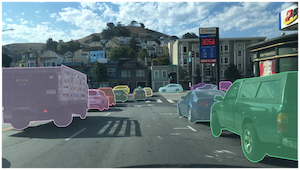}
        \\
       \includegraphics[width=0.19\textwidth]{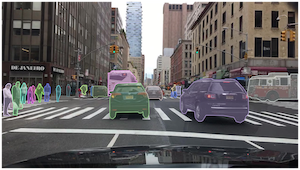} &
        \includegraphics[width=0.19\textwidth]{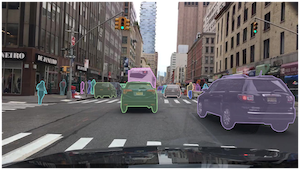} &
        \includegraphics[width=0.19\textwidth]{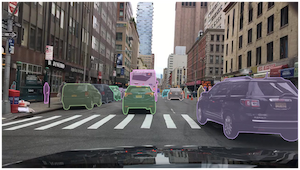} &
        \includegraphics[width=0.19\textwidth]{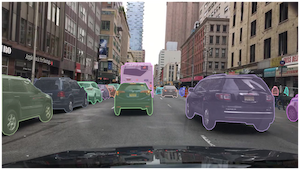} &
        \includegraphics[width=0.19\textwidth]{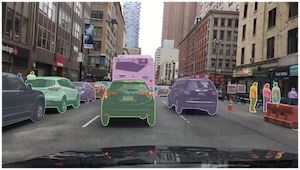}
        \\
        \includegraphics[width=0.19\textwidth]{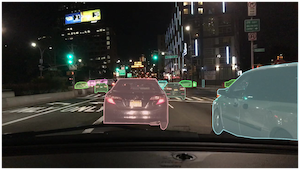} &
        \includegraphics[width=0.19\textwidth]{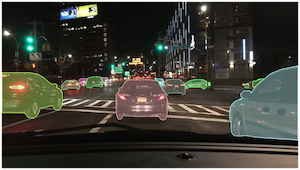} &
        \includegraphics[width=0.19\textwidth]{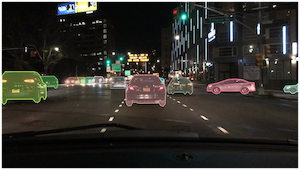} &
        \includegraphics[width=0.19\textwidth]{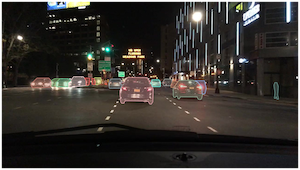} &
        \includegraphics[width=0.19\textwidth]{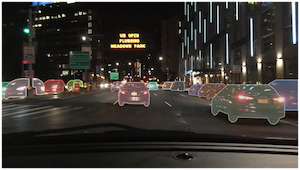}
        \\
        \includegraphics[width=0.19\textwidth]{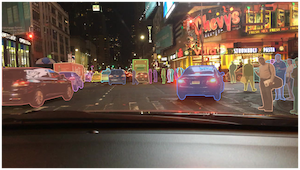} &
        \includegraphics[width=0.19\textwidth]{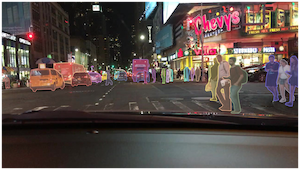} &
        \includegraphics[width=0.19\textwidth]{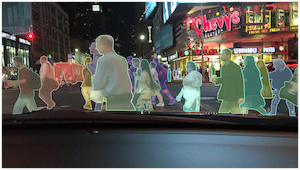} &
        \includegraphics[width=0.19\textwidth]{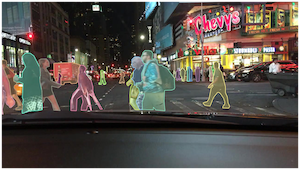} &
        \includegraphics[width=0.19\textwidth]{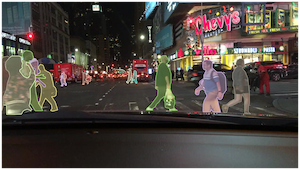}
        \\
    \end{tabular}
    \vspace{-2mm}
    \caption{\small Example annotations for BDD100K MOTS. Frames are down-sampled for visualization.}
    \label{fig:seg_tracking_label}
\end{figure*}

\begin{table*}[ht]
\centering
\adjustbox{width=.6\linewidth}{
\begin{tabular}{l|c|cccccc}
\toprule
Train (30K) & Test                         & AP            & AP$_{50}$     & AP$_{75}$     & AP$_S$        & AP$_M$        & AP$_L$        \\\midrule
City        & \multirow{3}{*}{City}        & \textbf{29.5} & \textbf{55.3} & \textbf{27.2} & \textbf{14.4} & \textbf{32.8} & \textbf{47.2} \\
Non-City    &                              & 24.9          & 48.6          & 22.1          & 11.7          & 28.1          & 40.7          \\
Random      &                              & 28.7          & 54.5          & 25.8          & 13.7          & 31.9          & 47.0          \\ \midrule
City        & \multirow{3}{*}{Non-City}    & 26.5          & 49.3          & \textbf{25.5} & 13.5          & \textbf{32.1} & 47.0          \\
Non-City    &                              & 24.3          & 46.0          & 22.4          & 13.3          & 30.0          & 42.0          \\
Random      &                              & \textbf{26.6} & \textbf{49.8} & 24.4          & \textbf{14.4} & 31.8          & \textbf{47.4} \\\midrule
City        & \multirow{3}{*}{Val}         & \textbf{28.8} & 54.1          & \textbf{26.8} & \textbf{13.8} & \textbf{32.7} & \textbf{47.0} \\
Non-City    &                              & 24.9          & 48.3          & 22.2          & 11.8          & 28.7          & 41.2          \\
Random      &                              & 28.7          & \textbf{54.5} & 25.8          & 13.7          & 31.9          & \textbf{47.0} \\\midrule
Daytime     & \multirow{3}{*}{Daytime}     & \textbf{30.6} & \textbf{56.0} & \textbf{28.7} & \textbf{16.4} & \textbf{35.6} & \textbf{50.7} \\
Non-Daytime &                              & 25.9          & 49.6          & 23.2          & 12.8          & 30.7          & 42.7          \\
Random      &                              & 29.5          & 55.0          & 27.2          & 15.7          & 34.5          & 48.7          \\\midrule
Daytime     & \multirow{3}{*}{Non-Daytime} & 23.6          & 46.1          & 21.2          & 10.5          & 25.4          & 41.3          \\
Non-Daytime &                              & 25.3          & 49.9          & 22.0          & 11.6          & 26.6          & 43.4          \\
Random      &                              & \textbf{26.0} & \textbf{50.9} & \textbf{22.5} & \textbf{12.8} & \textbf{27.4} & \textbf{45.1} \\\midrule
Daytime     & \multirow{3}{*}{Val}         & 28.1          & 52.8          & \textbf{25.9} & 13.3          & \textbf{31.9} & \textbf{47.0} \\
Non-Daytime &                              & 25.6          & 49.8          & 22.6          & 11.5          & 29.1          & 42.5          \\
Random      &                              & \textbf{28.7} & \textbf{54.5} & 25.8          & \textbf{13.7} & \textbf{31.9} & \textbf{47.0} \\
\bottomrule
\end{tabular}}
\caption{\small Full evaluation results of the domain discrepancy experiments with object detection.}

\end{table*}

\begin{table*}[htp]
\centering
\adjustbox{width=\linewidth}{
\begin{tabular}{c|l|ccc|ccc|ccccccc|c}
\toprule
\multirow{2}{*}{Threshold} & \multirow{2}{*}{Training Set} & \multicolumn{3}{c|}{Direction} & \multicolumn{3}{c|}{Continuity} & \multicolumn{7}{c|}{Category}                                                                  & \\
                           &                               & parallel & vertical & avg. & continuous  & dashed & avg. & {crosswalk} & \makecell{double \\ white} & \makecell{double\\ yellow} & \makecell{road\\ curb} & \makecell{single\\ white} & \makecell{single \\yellow} & {avg.} & \makecell{total \\ avg.}                          \\ \midrule
\multirow{6}{*}{$\tau=1$}  & Lane 10K                      & 28.41    & 28.35    & 28.38   & 28.31       & 26.32  & 27.31   & 27.48     & 6.5          & 32.99         & 19.92     & 28.51        & 27.09         & 23.75   & 26.48                          \\
                           & Lane+Drivable 10K             & 31.19    & 32.46    & 31.83   & 31.89       & 28.84  & 30.36   & 31.35     & 14.41        & 37            & 24.28     & 30.4         & 28.6          & 27.68   & 29.95                          \\
                           & Lane 20K                      & 34.45    & 36.62    & 35.54   & 34.58       & 33.61  & 34.09   & 35.73     & 20.75        & 39.7          & 27.59     & 34.53        & 33.5          & 31.97   & 33.87                          \\
                           & Lane+Drivable 20K             & 34.45    & 36.32    & 35.38   & 34.51       & 33.32  & 33.92   & 35.34     & 20.14        & 39.69         & 27.59     & 34.42        & 33.4          & 31.76   & 33.69                          \\
                           & Lane 70K                      & 34.57    & 36.92    & 35.74   & 34.62       & 33.85  & 34.23   & 36.17     & 21.51        & 39.88         & 27.91     & 34.62        & 33.77         & 32.31   & 34.1                           \\
                           & Lane+Drivable 70K             & 34.48    & 36.60    & 35.54   & 34.49       & 33.62  & 34.05   & 35.78     & 20.7         & 39.69         & 27.87     & 34.4         & 33.47         & 31.99   & 33.86                          \\ \midrule
\multirow{6}{*}{$\tau=2$}  & Lane 10K                      & 35.76    & 36.63    & 36.19   & 35.48       & 33.91  & 34.70   & 35.85     & 7.76         & 39.31         & 26.64     & 35.61        & 32.73         & 29.65   & 33.51                          \\
                           & Lane+Drivable 10K             & 38.79    & 41.26    & 40.03   & 39.28       & 37.01  & 38.14   & 40.26     & 16.94        & 43.34         & 31.78     & 37.78        & 34.78         & 34.15   & 37.44                          \\
                           & Lane 20K                      & 42.44    & 46.03    & 44.23   & 42.32       & 42.41  & 42.37   & 45.31     & 24.89        & 46.35         & 35.76     & 42.41        & 40.34         & 39.18   & 41.93                          \\
                           & Lane+Drivable 20K             & 42.42    & 45.65    & 44.03   & 42.22       & 42.06  & 42.14   & 44.78     & 24.07        & 46.38         & 35.77     & 42.23        & 39.99         & 38.87   & 41.68                          \\
                           & Lane 70K                      & 42.56    & 46.40    & 44.48   & 42.32       & 42.71  & 42.51   & 45.8      & 25.44        & 46.54         & 36.09     & 42.48        & 40.47         & 39.47   & 42.15                          \\
                           & Lane+Drivable 70K             & 42.48    & 46.00    & 44.24   & 42.18       & 42.46  & 42.32   & 45.32     & 24.6         & 46.39         & 36.08     & 42.25        & 40.09         & 39.12   & 41.89                          \\ \midrule
\multirow{6}{*}{$\tau=10$} & Lane 10K                      & 49.35    & 49.22    & 49.29   & 48.32       & 47.39  & 47.85   & 49.25     & 9.37         & 46.62         & 44.14     & 46.41        & 38.72         & 39.08   & 45.41                          \\
                           & Lane+Drivable 10K             & 54.07    & 53.87    & 53.97   & 52.61       & 52.57  & 52.59   & 53.37     & 20.64        & 51.05         & 50.27     & 50.58        & 41.98         & 44.65   & 50.4                           \\
                           & Lane 20K                      & 56.34    & 58.38    & 57.36   & 54.71       & 56.99  & 55.85   & 58.68     & 30.71        & 54.48         & 52.73     & 54.49        & 48.19         & 49.88   & 54.36                          \\
                           & Lane+Drivable 20K             & 56.31    & 58.07    & 57.19   & 54.59       & 56.69  & 55.64   & 58.24     & 29.39        & 54.68         & 52.86     & 54.22        & 47.62         & 49.5    & 54.11                          \\
                           & Lane 70K                      & 56.3     & 58.70    & 57.50   & 54.59       & 57.16  & 55.87   & 59.1      & 30.87        & 54.86         & 53.05     & 54.34        & 48.28         & 50.08   & 54.48                          \\
                           & Lane+Drivable 70K             & 56.41    & 58.29    & 57.35   & 54.53       & 56.98  & 55.76   & 58.53     & 29.63        & 54.6          & 53.06     & 54.22        & 47.72         & 49.63   & 54.24          \\ \bottomrule                
\end{tabular}}
\caption{\small Full evaluation results of the individual lane marking task and the joint training of lane marking and the drivable area detection. We report the ODS-F scores with different thresholds $\tau=1,2,10$ pixels of direction, continuity as well as each category. }
\label{tab:lane_eval}
\end{table*}

\section{Model Details}
In this section, we present more implementation details for benchmark models.

\subsection{Tracking}

We use a modified Faster R-CNN~\cite{ren2015faster} architecture for tracking similar with Feichtenhofer~\etal~\cite{d&t}. Like Feichtenhofer~\etal~\cite{d&t}, we use a correlation module and a bounding box propagation (regression) head to estimate the bounding box offset between two frames for short-term association. We also implement an association head based on appearance to learn embeddings for instance re-identification. During training, we sample a pair of frames within the interval of $t = 3$ frames. During inference, we first perform detection for the first frame. For each subsequent frame, we use the propagation head to associate detected bounding boxes with boxes from the previous frame based on overlap. We then use the association head based on appearance to associate the rest with the unmatched boxes in the previous 15 frames using dot product of the embeddings followed by softmax.

\end{document}